%% file: content.tex
\documentclass{article}

\usepackage[final, nonatbib]{neurips_2024}

\usepackage{abbrv}
\usepackage[dvipsnames]{xcolor}
\usepackage{url}
\usepackage{tikz}
\usepackage{comment}
\usepackage{amsmath,amssymb}
\usepackage{colortbl}
\usepackage{epsfig}
\usepackage[skip=0pt]{caption}
\usepackage{soul}
\usepackage[utf8]{inputenc}
\usepackage{listings}
\usepackage[T1]{fontenc}
\usepackage{booktabs}
\usepackage{amsfonts}
\usepackage{nicefrac}
\usepackage{microtype}
\usepackage{algorithmic-mine}
\usepackage{algorithm-mine}
\usepackage{numprint}
\usepackage{siunitx}
\usepackage{etoolbox}
\usepackage{cancel}
\newcommand{\ubold}{\fontseries{b}\selectfont}
\robustify\ubold
\usepackage{wrapfig}
\usepackage[skip=2pt,font=scriptsize,labelfont=scriptsize]{subcaption}
\usepackage{pifont}
\usepackage{lineno}
\usepackage{array,multirow,graphicx}
\usepackage[para]{footmisc}

\usepackage[accsupp]{axessibility}
\usepackage[pagebackref,breaklinks,colorlinks]{hyperref}
\usepackage{pgfplots}
\pgfplotsset{compat=1.18}

\newcommand\shadetext[2][]{%
 \setbox0=\hbox{{#2}}%
 \tikz[baseline=0]\path [#1] \pgfextra{\rlap{\copy0}} (0,-\dp0) rectangle (\wd0,\ht0);%
}

\definecolor{keyword}{rgb}{.224,.451,.686}

\newcommand{\xmark}{\textcolor{BrickRed}{\ding{55}}}

\newcommand{\keyword}[1]{\textcolor{keyword}{#1}}
\newcommand{\keywordone}[1]{\textcolor{Red}{#1}}
\newcommand{\keywordtwo}[1]{\textcolor{Purple}{#1}}
\newcommand{\keywordtri}[1]{\textcolor{Green}{#1}}

\newcommand{\TAP}{\textcolor{OliveGreen}{\textbf{TAP-Vid}}}
\newcommand{\MOT}{\textcolor{OliveGreen}{\textbf{MOT}}}
\newcommand{\VOS}{\textcolor{blue}{\textbf{VOS}}}
\newcommand{\MOTS}{\textcolor{OliveGreen}{\textbf{MOTS}}}
\newcommand{\KITTI}{\textcolor{OliveGreen}{\textbf{KITTI}}}
\newcommand{\VIS}{\textcolor{OliveGreen}{\textbf{VIS}}}
\newcommand{\PT}{\textcolor{OliveGreen}{\textbf{PoseTrack}}}
\newcommand{\LSOT}{\textcolor{blue}{\textbf{LaSOT}}}
\newcommand{\GOT}{\textcolor{OliveGreen}{\textbf{GroOT}}}
\newcommand{\method}{\shadetext[left color=blue, right color=OliveGreen, middle color=teal, shading angle=30]{DINTR}}

\title{\method: Tracking via Diffusion-based Interpolation}

\author{Pha Nguyen$^{1}$, Ngan Le$^{1}$, Jackson Cothren$^{1}$, \textbf{Alper Yilmaz}$^{2}$, \textbf{Khoa Luu}$^{1}$ \\
\small $^{1}$University of Arkansas \quad 
\small $^{2}$Ohio State University\\
\tt\small $^{1}$\{panguyen, thile, jcothre, khoaluu\}@uark.edu \quad 
$^{2}$yilmaz.15@osu.edu
}

\begin{document}

\input{revision}

{\small
\bibliographystyle{unsrt}
\bibliography{egbib}
}

\newpage

\renewcommand{\thefigure}{\Alph{section}.\arabic{figure}}
\renewcommand{\thetable}{\Alph{section}.\arabic{table}}
\renewcommand{\theequation}{\Alph{section}.\arabic{equation}}
\renewcommand{\thealgorithm}{\Alph{section}.\arabic{algorithm}}

\input{revised_supp}

\clearpage
\newpage

\end{document}

%% file: revision.tex
\maketitle

\begin{center}
 \centering
 \vspace{-\baselineskip}
 \captionsetup{type=figure}
 \includegraphics[width=\textwidth]{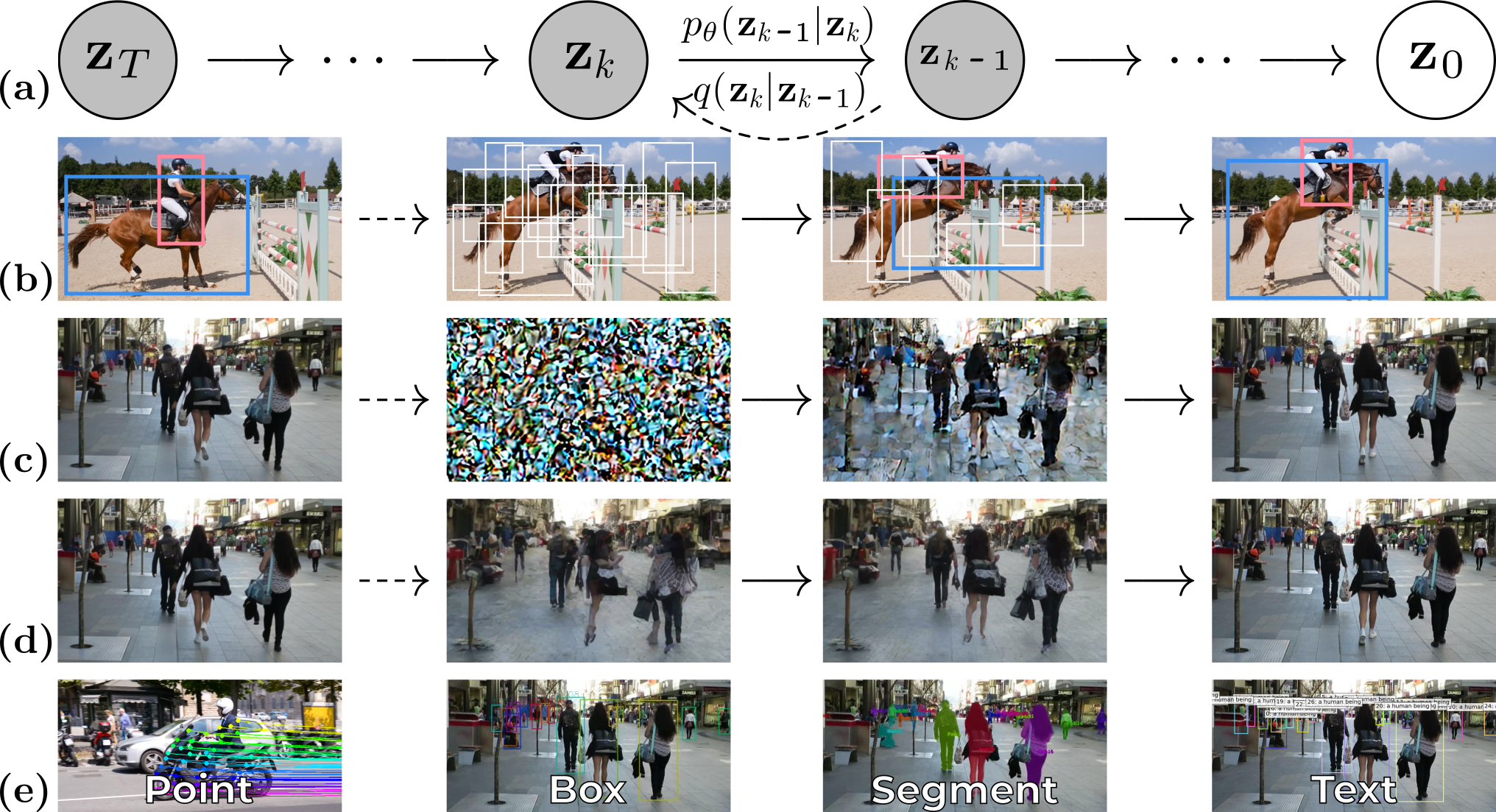}
 \captionof{figure}{Diffusion-based processes. (a) Probabilistic diffusion process~\cite{ho2020denoising}, where $q(\cdot)$ is noise sampling and $p_\theta(\cdot)$ is denoising. (b) Diffusion process in the 2D coordinate space~\cite{chen2023diffusiondet, luo2023diffusiontrack, lv2024diffmot}. (c) A purely visual diffusion-based \textit{data prediction} approach reconstructs the subsequent video frame. (d) Our proposed \textit{data interpolation} approach \textbf{\method} interpolates between two consecutive video frames, indexed by timestamp $t$, allowing a seamless temporal transition for visual content understanding, temporal modeling, and instance extracting for the object tracking task across various indications (e).
 }
 \label{fig:teaser}
\end{center}

\begin{abstract}

Object tracking is a fundamental task in computer vision, requiring the localization of objects of interest across video frames. Diffusion models have shown remarkable capabilities in visual generation, making them well-suited for addressing several requirements of the tracking problem. This work proposes a novel diffusion-based methodology to formulate the tracking task. Firstly, their conditional process allows for injecting indications of the target object into the generation process. Secondly, diffusion mechanics can be developed to inherently model temporal correspondences, enabling the reconstruction of actual frames in video. However, existing diffusion models rely on extensive and unnecessary mapping to a Gaussian noise domain, which can be replaced by a more efficient and stable interpolation process. Our proposed interpolation mechanism draws inspiration from classic image-processing techniques, offering a more interpretable, stable, and faster approach tailored specifically for the object tracking task. By leveraging the strengths of diffusion models while circumventing their limitations, our \shadetext[left color=blue, right color=OliveGreen, middle color=teal, shading angle=30]{\textbf{D}}iffusion-based \shadetext[left color=blue, right color=OliveGreen, middle color=teal, shading angle=30]{\textbf{IN}}terpolation \shadetext[left color=blue, right color=OliveGreen, middle color=teal, shading angle=30]{\textbf{T}}racke\shadetext[left color=blue, right color=OliveGreen, middle color=teal, shading angle=30]{\textbf{R}} (\textbf{\method}) presents a promising new paradigm and achieves a superior multiplicity on seven benchmarks across five indicator representations.

\end{abstract}

\section{Introduction}\label{sec:intro}

Object tracking is a long-standing computer vision task with widespread applications in video analysis and instance-based understanding. Over the past decades, numerous tracking paradigms have been explored, including \textit{tracking-by-regression}~\cite{zhou2020tracking}, \textit{-detection}~\cite{wojke2017simple}, \textit{-segmentation}~\cite{wu2021track} and two more recent \textit{tracking-by-attention}~\cite{meinhardt2022trackformer, zeng2022motr}, \textit{-unification}~\cite{yan2022towards} paradigms. Recently, generative modeling has achieved great success, offering several promising new perspectives in instance recognition. These include denoising sampling bounding boxes to final prediction~\cite{chen2023diffusiondet, luo2023diffusiontrack, lv2024diffmot}, or sampling future trajectories~\cite{dendorfer2022quo}. Although these studies explore the generative process in instance-based understanding tasks, they perform solely on coordinate refinement rather than performing on the visual domain, as in Fig.~\ref{fig:teaser}b.

In this work, we propose a novel tracking framework solely based on \textit{visual} iterative latent variables of diffusion models~\cite{sohl2015deep, rombach2022high}, thereby introducing the novel and true \textit{Tracking-by-Diffusion} paradigm. This paradigm demonstrates versatile applications across various indications, comprising points, bounding boxes, segments, and textual prompts, facilitated by the conditional mechanism (Eqn.~\eqref{eq:attn}).

Moreover, our proposed \shadetext[left color=blue, right color=OliveGreen, middle color=teal, shading angle=30]{\textbf{D}}iffusion-based \shadetext[left color=blue, right color=OliveGreen, middle color=teal, shading angle=30]{\textbf{IN}}terpolation \shadetext[left color=blue, right color=OliveGreen, middle color=teal, shading angle=30]{\textbf{T}}racke\shadetext[left color=blue, right color=OliveGreen, middle color=teal, shading angle=30]{\textbf{R}} (\textbf{\method}) inherently models the temporal correspondences via the diffusion mechanics, \ie, the denoising process. Specifically, by formulating the process to operate temporal modeling \textit{online} and \textit{auto-regressively} (\ie next-frame reconstruction, as in Eqn.~\eqref{eq:prediction}), \textbf{\method} enables the capability for instance-based video understanding tasks, specifically the object tracking. However, existing diffusion mechanics rely on an extensive and unnecessary mapping to a Gaussian noise domain, which we argue can be replaced by a more efficient interpolation process (Subsection~\ref{method}). Our proposed interpolation operator draws inspiration from the image processing field, offering a more direct, seamless, and stable approach. By leveraging the diffusion mechanics while circumventing their limitations, our \textbf{\method} achieves superior multiplicity on seven benchmarks across five types of indication, as elaborated in Section~\ref{sec:exp}. Note that our Interpolation process does not aim to generate high-fidelity unseen frames~\cite{reda2022film, huang2022real, jain2024video, li2023amt}. Instead, its objective is to seamlessly transfer internal states between frames for visual semantic understanding.

\textbf{Contributions.} Overall, \textit{\textbf{(i)}} this paper reformulates the \textit{Tracking-by-Diffusion} paradigm to operate on visual domain \textit{\textbf{(ii)}} which demonstrates broader tracking applications than existing paradigms. \textit{\textbf{(iii)}} We reformulate the diffusion mechanics to achieve two goals, including \textit{(a)} temporal modeling and \textit{(b)} iterative interpolation as a $2\times$ faster process. \textit{\textbf{(iv)}} Our proposed \textbf{\method} achieves superior multiplicity and State-of-the-Art (SOTA) performances on \textit{seven tracking benchmarks} of \textit{five representations}. \textit{\textbf{(v)}} Following sections including \textbf{Appendices}~\ref{appendix} elaborate on its formulations, properties, and evaluations.

\section{Related Work}\label{sec:related}

\subsection{Object Tracking Paradigms}

\begin{table*}[!t]
 \caption{Comparison of paradigms, mechanisms of SOTA tracking methods. \textbf{Indication Types} defines the representation to indicate targets with their corresponding datasets: \TAP~\cite{doersch2022tap}, \PT~\cite{andriluka2018posetrack,doering2022posetrack21}, \MOT~\cite{MOTChallenge2015, MOT16, sun2022dancetrack}, \VOS~\cite{perazzi2016benchmark}, \VIS~\cite{yang2019video}, \MOTS~\cite{porzi2020learning}, \KITTI~\cite{geiger2012we}, \LSOT~\cite{fan2021lasot}, \GOT~\cite{nguyen2023type}. \shadetext[left color=blue, right color=OliveGreen, middle color=teal, shading angle=30]{\textbf{Methods}} in color gradient support both types of \textcolor{blue}{\textbf{single-}} and \textcolor{OliveGreen}{\textbf{multi-}}target benchmarks.}\label{tab:discussion}
 \resizebox{\textwidth}{!}{
 \setlength\tabcolsep{2pt}
 \begin{tabular}{lccccccc}
 \toprule
 \multicolumn{1}{c}{\multirow{2}{*}{\textbf{Method}}} & \multicolumn{1}{c}{\multirow{2}{*}{\textbf{Paradigm}}} & \multicolumn{1}{c}{\multirow{2}{*}{\textbf{Mechanism}$^*$}} & \multicolumn{5}{c}{\textbf{Indication Types}} \\ \cmidrule{4-8}
 & & & Point & Pose & Box & Segment & Text \\
 \midrule
 TAPIR~\cite{doersch2023tapir} & \multirow{4}{*}{Regression} & \textit{Iter.} Refinement & \TAP & \xmark & \xmark & \xmark & \xmark \\
 Tracktor++~\cite{bergmann2019tracking} & & Regression Head & \xmark & \xmark & \MOT & \xmark & \xmark \\
 CenterTrack~\cite{zhou2020tracking} & & Offset Prediction & \xmark & \xmark & \MOT & \xmark & \xmark \\
 GTI~\cite{yang2020grounding} & & \textit{Rgn-Tpl Integ.} & \xmark & \xmark & \LSOT & \xmark & \LSOT \\ \midrule
 DeepSORT~\cite{wojke2018deep} & \multirow{4}{*}{Detection} & Cascade \textit{Assoc.} & \xmark & \xmark & \MOT & \xmark & \xmark \\
 GSDT~\cite{wang2021joint} & & Relation Graph & \xmark & \xmark & \MOT & \xmark & \xmark \\
 JDE~\cite{wang2019towards} & & Multi-Task & \xmark & \xmark & \MOT & \xmark & \xmark \\
 ByteTrack~\cite{zhang2021bytetrack} & & Two-stage \textit{Assoc.} & \xmark & \xmark & \MOT & \xmark & \xmark \\ \midrule
 TrackR-CNN~\cite{voigtlaender2019mots} & \multirow{4}{*}{Segmentation} & 3D Convolution & \xmark & \xmark & \xmark & \MOTS & \xmark \\
 MOTSNet~\cite{porzi2020learning} & & Mask-Pooling & \xmark & \xmark & \xmark & \MOTS & \xmark \\
 CAMOT~\cite{ovsep2018track} & & Hypothesis Select & \xmark & \xmark & \xmark & \KITTI & \xmark \\
 PointTrack~\cite{xu2020segment} & & \textit{Seg.} as Points & \xmark & \xmark & \xmark & \MOTS/\KITTI & \xmark \\ \midrule
 MixFormerV2~\cite{cui2024mixformerv2} & \multirow{4}{*}{Attention} & Mixed Attention & \xmark & \xmark & \LSOT & \xmark & \xmark \\
 TransVLT~\cite{zhao2023transformer} & & \textit{X}-Modal Fusion & \xmark & \xmark & \LSOT & \xmark & \LSOT \\
 MeMOTR~\cite{gao2023memotr} & & Memory \textit{Aug.} & \xmark & \xmark & \MOT & \xmark & \xmark \\
 MENDER~\cite{nguyen2023type} & & Tensor \textit{Decomp.} & \xmark & \xmark & \MOT & \xmark & \GOT \\ \midrule
 SiamMask~\cite{wang2019fast} & \multirow{4}{*}{Unification} & Variant Head & \xmark & \xmark & \LSOT & \VOS & \xmark \\
 TraDeS~\cite{wu2021track} & & Cost Volume & \xmark & \xmark & \MOT & \VIS/\MOTS & \xmark \\
 \shadetext[left color=blue, right color=OliveGreen, middle color=teal, shading angle=30]{UNICORN}~\cite{yan2022towards} & & Unified \textit{Embed.} & \xmark & \xmark & \LSOT/\MOT & \VOS/\MOTS & \xmark \\
 \shadetext[left color=blue, right color=OliveGreen, middle color=teal, shading angle=30]{UniTrack}~\cite{wang2021different} & & Primitive Level & \xmark & \PT & \LSOT/\MOT & \VOS/\MOTS & \xmark \\ \midrule
 DiffusionTrack~\cite{luo2023diffusiontrack} & \multirow{3}{*}{\textbf{Diffusion}} & Denoised \textcolor{BrickRed}{\textit{Coord.}} & \xmark & \xmark & \MOT & \xmark & \xmark \\
 DiffMOT~\cite{lv2024diffmot} & & \textcolor{BrickRed}{\textit{Motion}} Predictor & \xmark & \xmark & \MOT & \xmark & \xmark \\
 \textbf{\method} \textbf{(Ours)} & & \textcolor{OliveGreen}{Visual} \textit{Interpolat.} & \TAP & \PT & \LSOT/\MOT & \VOS/\MOTS & \LSOT/\GOT \\
 \bottomrule
 \end{tabular}
 }
 {\small $^*$ \textit{Iter.}: Iterative. \textit{Rgn-Tpl Integ.}: Region-Template Integration. \textit{Assoc.}: Association. \textit{X}: Cross. \textit{Decomp.}: Decomposition. \textit{Embed.}: Embedding. \textcolor{BrickRed}{\textit{Coord.}}: \textcolor{BrickRed}{\textbf{2D} Coordinate}. \textcolor{BrickRed}{\textit{Motion}}: \textcolor{BrickRed}{\textbf{2D} Motion}. \textit{Interpolat.}: Interpolation.}
\end{table*}

\noindent\textbf{Tracking-by-Regression} methods refine future object positions directly based on visual features. Previous approaches~\cite{bergmann2019tracking, feichtenhofer2017detect} rely on the regression branch of object features in nearby regions. CenterTrack~\cite{zhou2020tracking} represents objects via center points and temporal offsets. It lacks explicit object identity, requiring the appearance~\cite{bergmann2019tracking}, motion model~\cite{liu2020gsm}, and graph matching~\cite{braso2020learning} components.

\noindent\textbf{Tracking-by-Detection} methods form object trajectories by linking detections over consecutive frames, treating the task as an optimization problem.
\textit{Graph}-based methods formulate the tracking problem as a bipartite matching or maximum flow~\cite{berclaz2011multiple}. These methods utilize a variety of techniques, such as link prediction~\cite{quach2021dyglip}, trainable graph neural networks~\cite{braso2020learning, wang2021joint}, edge lifting~\cite{hornakova2020lifted}, weighted graph labeling~\cite{henschel2017improvements}, multi-cuts~\cite{tang2017multiple, nguyen2022lmgp}, general-purpose solvers~\cite{yu2007multiple}, motion information~\cite{keuper2018motion}, learned models~\cite{kim2015multiple}, association graphs~\cite{sheng2018heterogeneous}, and distance-based~\cite{jiang2007linear, pirsiavash2011globally, zhang2008global}.
Additionally, \textit{Appearance}-based methods leverage robust image recognition frameworks to track objects. These techniques depend on similarity measures derived from 3D appearance and pose~\cite{rajasegaran2022tracking}, affinity estimation~\cite{chu2019famnet}, detection candidate selection~\cite{chu2019famnet}, learned re-identification features~\cite{pang2021quasi, ristani2018features}, or twin neural networks~\cite{leal2016learning}.
On the other hand, \textit{Motion} modeling is leveraged for camera motion~\cite{aharon2022bot}, observation-centric manner~\cite{cao2023observation}, trajectory forecasting~\cite{dendorfer2022quo}, the social force model~\cite{leal2011everybody, pellegrini2009you, scovanner2009learning, yamaguchi2011you}, based on constant velocity assumptions~\cite{andriyenko2011multi, chen2018real}, or location estimation~\cite{alahi2016social, leal2011everybody, robicquet2016learning} directly from trajectory sequences. Additionally, data-driven motion~\cite{leal2014learning} need to project 3D into 2D motions~\cite{tokmakov2021learning}.

\noindent\textbf{Tracking-by-Segmentation} leverages detailed pixel information and addresses the challenges of unclear backgrounds and crowded scenes. Methods include cost volumes~\cite{wu2021track}, point cloud representations~\cite{xu2020segment}, mask pooling layers~\cite{porzi2020learning}, and mask-based~\cite{ovsep2018track} with 3D convolutions~\cite{voigtlaender2019mots}. However, its reliance on segmented multiple object tracking data often necessitates bounding box initialization.

\noindent\textbf{Tracking-by-Attention} applies the attention mechanism~\cite{vaswani2017attention} to link detections with tracks at the feature level, represented as tokens. TrackFormer~\cite{meinhardt2022trackformer} approaches tracking as a unified prediction task using attention, during initiation. MOTR~\cite{zeng2022motr} and MOTRv2~\cite{zhang2023motrv2} advance this concept by integrating motion and appearance models, aiding in managing object entrances/exits and temporal relations.
Furthermore, object token representations can be enhanced via memory techniques, such as memory augmentation~\cite{gao2023memotr} and memory buffer~\cite{cai2022memot, zhou2024reading}. Recently, MENDER~\cite{nguyen2023type} presents another stride, a transformer architecture with tensor decomposition to facilitate object tracking through descriptions.

\noindent\textbf{Tracking-by-Unification} aims to develop unified frameworks that can handle multiple tasks simultaneously. Pioneering works in this area include TraDeS~\cite{wu2021track} and SiamMask~\cite{wang2019fast}, which combine object tracking (SOT/MOT) and video segmentation (VOS/VIS). UniTrack~\cite{wang2021different} employs separate task-specific heads, enabling both object propagation and association across frames. Furthermore, UNICORN~\cite{yan2022towards} investigates learning robust representations by consolidating from diverse datasets.

\subsection{Diffusion Model in Semantic Understanding}

Generative models have recently been found to be capable of performing understanding tasks.

\noindent\textbf{Visual Representation and Correspondence.} Hedlin \etal~\cite{hedlin2024unsupervised} establishes semantic visual correspondences by optimizing text embeddings to focus on specific regions.
Diffusion Autoencoders~\cite{preechakul2022diffusion} form a diffusion-based autoencoder encapsulating high-level semantic information.
Similarly, Zhang \etal~\cite{zhang2023tale} combine features from Stable Diffusion (SD) and DINOv2~\cite{oquab2023dinov2} models, effectively merging the high-quality spatial information and capitalizing on both strengths. 
Diffusion Hyperfeatures~\cite{luo2023dhf} uses feature aggregation and transforms intermediate feature maps from the diffusion process into a single, coherent descriptor map.
Concurrently, DIFT~\cite{tang2023emergent} simulates the forward diffusion process, adding noise to input images and extracting features within the U-Net.
Asyrp~\cite{kwon2023diffusion} employs the asymmetric reverse process to explore and manipulate a semantic latent space, upholding the original performance, integrity, and consistency.
Furthermore, DRL~\cite{mittal2023diffusion} introduces an infinite-dimensional latent code that offers discretionary control over the granularity of detail.

\noindent\textbf{Generative Perspectives in Object Tracking.} A straightforward application of generative models in object tracking is to augment and enrich training data~\cite{prakash2021gan, li2023trackdiffusion, nguyen2024dataset}.
For trajectory refinement, QuoVadis~\cite{dendorfer2022quo} uses the social generative adversarial network (GAN)~\cite{gupta2018social} to sample future trajectories to account for the uncertainty in future positions.
DiffusionTrack~\cite{luo2023diffusiontrack} and DiffMOT~\cite{lv2024diffmot} utilize the diffusion process in the bounding box decoder. Specifically, they pad prior \textit{2D coordinate} bounding boxes with noise, then transform them into tracking results via a denoising decoder.

\subsection{Discussion}\label{subsec:discuss}
This subsection discusses the key aspects of our proposed paradigm and method, including the mechanism comparison of our \textbf{\method} against alternative diffusion approaches~\cite{chen2023diffusiondet, luo2023diffusiontrack, lv2024diffmot}, and the properties that enable \textit{Tracking-by-Diffusion} on visual domain to stand out from the existing paradigms.

\noindent\textbf{Conditioning Mechanism.} As illustrated in Fig.~\ref{fig:teaser}b, tracking methods performing diffusion on the 2D coordinate space~\cite{luo2023diffusiontrack, lv2024diffmot} utilize generative models to model 2D object motion or refine coordinate predictions. However, they fail to leverage the conditioning mechanism~\cite{rombach2022high} of Latent Diffusion Models, which are principally capable of modeling unified conditional distributions. As a result, these diffusion-based approaches have a specified indicator representation limited to the bounding box, that cannot be expanded to other advanced indications, such as point, pose, segment, and text.

In contrast, we formulate the object tracking task as two visual processes, including one for diffusion-based Reconstruction, as illustrated in Fig.~\ref{fig:teaser}c, and another $2\times$ faster approach that is Interpolation, as shown in Fig.~\ref{fig:teaser}d. These two approaches demonstrate their superior versatility due to the controlled injection $p_\theta(\mathbf{z} | \tau)$ implemented by the attention mechanism~\cite{vaswani2017attention} (Eqn.~\eqref{eq:attn}) during iterative diffusion.

\noindent\textbf{Unification.} Current methods under \textit{tracking-by-unification} face challenges due to the separation of task-specific heads. This issue arises because single-object and multi-object tracking tasks are trained on distinct branches~\cite{wu2021track, wang2021different} or stages~\cite{wang2019towards}, with results produced through a manually designed decoder for each task. The architectural discrepancies limit the full utilization of network capacity.

In contrast, \textit{Tracking-by-Diffusion} operating on the visual domain addresses the limitations of unification. Our method seamlessly handles diverse tracking objectives, including \textit{(a) point and pose \textbf{regression}}, \textit{(b) bounding box and segmentation \textbf{prediction}}, and \textit{(c) referring \textbf{initialization}}, while remaining \textit{(d) \textbf{data-} and \textbf{process-unified}} through an iterative process. This is possible because our approach operates on the base core domain, allowing it to understand contexts and extract predictions.

\noindent\textbf{Application Coverage} presented in Table~\ref{tab:discussion} validates the unification advantages of our approach. As highlighted, our proposed model \textbf{\method} supports unified tracking across \textit{seven benchmarks} of \textit{eight settings} comprising \textit{five distinct categories of indication}. It can handle both \textcolor{blue}{\textbf{single-target}} and \textcolor{OliveGreen}{\textbf{multiple-target}} benchmarks, setting a new standard in terms of multiplicity, flexibility, and novelty.

\section{Problem Formulation}

Given two images $\mathbf{I}_t$ and $\mathbf{I}_{t+1}$ from a video sequence $\mathcal{V}$, and an indicator representation $L_t$ (\eg, point, structured points set for pose, bounding box, segment, or text) for an object in $\mathbf{I}_t$, our goal is to find the respective region $L_{t+1}$ in $\mathbf{I}_{t+1}$. The relationship between $L_t$ and $L_{t+1}$ can encode semantic correspondences~\cite{tang2023emergent, luo2023dhf, namekata2024emerdiff} (\ie, different objects with similar semantic meanings), geometric correspondence~\cite{nguyen2022multi, nguyen2023utopia, truong2023crovia} (\ie, the same object viewed from different viewpoints) or temporal correspondence~\cite{karaev2023cotracker, moing2023dense, xiao2024spatialtracker} (\ie, the location of a deforming object over a video sequence).

We define the object-tracking task as temporal correspondence, aiming to establish matches between regions representing the same real-world object as it moves, potentially deforming or occluding across the video sequence over time. Let us denote a feature encoder $\mathcal{E}(\cdot)$ that takes as input the frame $\mathbf{I}_t$ and returns the feature representation $\mathbf{z}^t$. Along with the region $L_t$ for the initial indication, the \textit{\textbf{online} and \textbf{auto-regressive} objective} for the tracking task can be written as follows:
\begin{equation}\label{temporal_objective}
 \small
 \keyword{L_{t+1}} = \mathrm{arg}\min_{\keyword{L}} dist\big(\mathcal{E}(\mathbf{I}_t) [L_t], \mathcal{E}(\mathbf{I}_{t+1})[\keyword{L}]\big),
\end{equation}
where $dist(\cdot, \cdot)$ is a semantic distance that can be cosine~\cite{wojke2018deep} or distributional softmax~\cite{fischer2023qdtrack}. A special case is to give $L_t$ as textual input and return $L_{t+1}$ as a bounding box for the \textit{referring object tracking}~\cite{nguyen2023type, wu2023referring} task. In addition, the pose is treated as multiple-point tracking. The output $\keyword{L_{t+1}}$ is then mapped to a point, box, or segment. We explore how diffusion models can learn these temporal dynamics end-to-end to output consistent object representations frame-to-frame in the next section.

\section{Methodology}

This section first presents the notations and background. Then, we present the deterministic frame reconstruction task for video modeling. Finally, our proposed framework \textbf{\method} is introduced.

\subsection{Notations and Background}\label{sec:preliminaries}

\textbf{Latent Diffusion Models} (LDMs)~\cite{ho2020denoising, rombach2022high, song2020denoising} are introduced to denoise the latent space of an autoencoder. First, the encoder $\mathcal{E}(\cdot)$ compresses a RGB image $\mathbf{I}_t$ into an initial latent space $\mathbf{z}^t_0=\mathcal{E}(\mathbf{I}_t)$, which can be reconstructed to a new image $\mathcal{D}(\mathbf{z}^t_0)$.
Let us denote two operators $\mathcal{Q}$ and $\mathcal{P}_{\varepsilon_\theta}$ are corresponding to the sampling noise process $q(\mathbf{z}^t_k | \mathbf{z}^t_{k - 1})$ and the denoising process $p_\varepsilon(\mathbf{z}^t_{k-1} | \mathbf{z}^t_{k})$, where $\mathcal{P}_{\varepsilon_\theta}$ is parameterized by an U-Net $\varepsilon_\theta$~\cite{ronneberger2015u} as a \textit{\keywordone{noise} prediction model} via the objective:
\begin{equation}
 \label{eq:diffusion}
 \small
 \min_\theta \mathbb{E}_{\mathbf{z}^t_0, \keywordone{\epsilon \sim \mathcal{N}(0, 1)}, \keywordtwo{k \sim \mathcal{U}(1, T)}}\Big[\big\|\keywordone{\epsilon} - \mathcal{P}_{\varepsilon_\theta}\big(\mathcal{Q}(\mathbf{z}^t_0, \keywordtwo{k}), \keywordtwo{k}, \tau\big)\big \|_2^2\Big], \qquad \text{where } \tau = \mathcal{T}_\theta(L_t).
\end{equation}

\textbf{Localization.} All types of localization $L_t$, \eg, point, pose (\ie set of structured points), bounding box, segment, and especially text, are unified as guided indicators. $\mathcal{T}_\theta(\cdot)$ is the respective extractor, such as the Gaussian kernel for point, pooling layer for bounding box and segment, or word embedding model for text. $\mathbf{z}^t_k$ is a noisy sample of $\mathbf{z}^t_0$ at step $k \in [1, \dots, T]$, and $T = 50$ is the maximum step.

\textbf{The Conditional Process $p_\theta(\mathbf{z}^{t+1}_0 | \tau)$}, containing cross-attention $Attn(\varepsilon, \tau)$ to inject the indication $\tau$ to an autoencoder with U-Net blocks $\varepsilon_\theta(\cdot, \cdot)$, is derived after noise sampling $\mathbf{z}^t_{k} = \mathcal{Q}(\mathbf{z}^t_{0}, k)$:
\begin{equation}
 \small
 \mathcal{P}_{\varepsilon_\theta}\big(\mathcal{Q}(\mathbf{z}^t_0, k), k, \tau\big) = \underbrace{\mathrm{softmax}\Big(\frac{\varepsilon_\theta(\overbrace{\sqrt{\bar{\alpha}_k}\mathbf{z}^t_{0} + \sqrt{1 - \bar{\alpha}_k}\epsilon}^{\mathcal{Q}(\mathbf{z}^t_{0}, k)})\times W_{Q}\times(\tau\times W_{K})^\intercal}{\sqrt{d}}\Big)}_{Attn(\varepsilon, \tau)}\times (\tau \times W_{V}),
 \label{eq:attn}
\end{equation}
where $W_{Q,K,V}$ are projection matrices, $d$ is the feature size, and $\alpha_k$ is a scheduling parameter.

\subsection{Deterministic Next-Frame Reconstruction by Data Prediction Model}

The \textit{noise prediction model}, defined in Eqn.~\eqref{eq:diffusion}, can not generate specific desired pixel content while denoising the latent feature to the new image. To effectively model and generate exactly the desired video content, we formulate a next-frame reconstruction task, such that $\mathcal{D}(\mathcal{P}_{\varepsilon_\theta}(\mathbf{z}^{t}_T, T, \tau)) \approx \mathbf{I}_{t+1}$. In this formulation, the denoised image obtained from the diffusion process should approximate the next frame in the video sequence. The objective for a \textit{\keywordtri{data} prediction model} (Fig.~\ref{fig:teaser}c) derives that goal as:
\begin{equation}
 \label{eq:prediction}
 \small
 \min_\theta \mathbb{E}_{\mathbf{z}^{t, \keywordtri{t+1}}_0, \keywordtwo{k \sim \mathcal{U}(1, T)}}\Big[\big\|\keywordtri{\mathbf{z}^{t+1}_k} - \mathcal{P}_{\varepsilon_\theta}\big(\mathcal{Q}(\mathbf{z}^{t}_0, \keywordtwo{k}), \keywordtwo{k}, \tau\big)\big \|_2^2\Big].
\end{equation}

\begin{wrapfigure}[10]{r}{0.58\textwidth}
 \begin{minipage}{0.58\textwidth}
 \vspace{-2\baselineskip}
 \begin{algorithm}[H] %
 \caption{Inplace Reconstruction Finetuning}
 \label{alg:finetuning}
 \begin{algorithmic}[1]
 \INPUT{Network $\varepsilon_\theta$, video sequence $\mathcal{V}$, indication $L_{t=0}$} %
 \STATE Sample $(t, t+1) \sim \mathcal{U}(0, |\mathcal{V}| - 2)$
 \STATE \hspace{0.5em} $\tau \gets \mathcal{T}_\theta(L_t)$
 \STATE \hspace{0.5em} Draw $\mathbf{I}_{t, t+1} \in \mathcal{V}$ and encode $\mathbf{z}^{t, t+1}_0 = \mathcal{E}(\mathbf{I}_{t, t+1})$
 \STATE \hspace{0.5em} Sample $k \sim \mathcal{U}(1,T)$
 \STATE \hspace{0.5em} \hspace{0.5em} Optimize $\min_\theta \big[\|\mathbf{z}^{t + 1}_k - \mathcal{P}_{\varepsilon_\theta}(\mathcal{Q}(\mathbf{z}^{t}_0, k), k, \tau)\|_2^2\big]$
 \STATE \hspace{0.5em} Optimize $\min_\theta \big[\|\mathbf{I}_{t + 1} - \mathcal{D}(\mathcal{P}_{\varepsilon_\theta}(\mathcal{Q}(\mathbf{z}^{t}_0, k), k, \tau))\|_2^2\big]$
 \end{algorithmic}
 \end{algorithm}
 \end{minipage}
\end{wrapfigure}

In layman's terms, the objective of the \textit{data prediction model} formulates the task of establishing temporal correspondence between frames by effectively capturing the pixel-level changes and reconstructing the real next frame from the current frame. With the pre-trained decoder $\mathcal{D}(\cdot)$ in place, the key optimization target becomes the denoising process itself.
To achieve this, a combination of \keywordtwo{step-wise} KL divergences is used to guide the likelihood of current frame latents $\mathbf{z}^{t}_k$ toward the desired latent representations for the next frame $\mathbf{z}^{t+1}_{k}$, as described in Alg.~\ref{alg:finetuning} and derived as:
\begin{equation}
 \small
 \mathcal{L} = \frac{1}{2}\mathbb{E}_{\mathbf{z}^{t, \keywordtri{t+1}}_0, \keywordtwo{k \sim \mathcal{U}(1, T)}} \big[\|\keywordtri{\mathbf{z}^{t+1}_k} - \mathcal{P}_{\varepsilon_\theta}(\mathcal{Q}(\mathbf{z}^{t}_0, \keywordtwo{k}), \keywordtwo{k}, \tau)\|_2^2\big] = \keywordtwo{\int_0^1}\frac{d}{d\keywordtwo{\alpha_k}} D_{KL}\big(q(\keywordtri{\mathbf{z}^{t+1}_{k}}|\keywordtri{\mathbf{z}^{t+1}_{k-1}}) \| p_\varepsilon(\mathbf{z}^{t}_{k-1} | \mathbf{z}^{t}_{k})\big)\,d\keywordtwo{\alpha_k}.\label{eq:data_prediction}
\end{equation}
where $\alpha_k = \frac{k}{T}$. This loss function constructed from the extensive \keywordtwo{step-wise} divergences creates an accumulative path between the visual distributions.
Instead, we propose to employ the classic interpolation operator used in image processing to formulate a new diffusion-based process that iteratively learns to blend video frames. This interpolation approach ultimately converges towards the same deterministic mapping toward $\mathbf{z}^{t+1}_0$ but is simpler to derive and more stable. The proposed process is illustrated in Fig.~\ref{fig:interpolation}, and interpolation operators are elaborated in the next Subsection~\ref{method}.

\subsection{\textbf{\method} for Tracking via Diffusion-based Interpolation}\label{method}

\textbf{Denoising Process as Temporal Interpolation.} We relax the controlled Gaussian space projection of \keywordtwo{every step}. Specifically, we impose a temporal bias by training a \textit{\keywordtri{data} \keyword{interpolation} model} $\keyword{\phi_\theta}$. The data interpolation process is denoted as $\mathcal{P}_{\keyword{\phi_\theta}}$ producing intermediate interpolated features $\widehat{\mathbf{z}}^{t+1}_k$, so that $\mathcal{P}_{\keyword{\phi_\theta}}(\mathbf{z}^{t}_0, T, \tau) = \widehat{\mathbf{z}}^{t+1}_0 \approx \mathbf{z}^{t+1}_0$. The goal is to obtain $p_{\keyword{\phi}}(\mathbf{z}^{t+1}_0|\mathbf{z}^{t}_0)$ by optimizing the objective:
\begin{equation}
 \label{eq:interpolation}
 \small
 \min_\theta \mathbb{E}_{\mathbf{z}^{t, \keywordtri{t+1}}_0}\big[\|\keywordtri{\mathbf{z}^{t + 1}_0} - \mathcal{P}_{\keyword{\phi_\theta}}(\mathbf{z}^{t}_0, T, \tau) \|_2^2\big].
\end{equation}

\begin{figure}[!t]
 \begin{minipage}{0.465\textwidth}
 \resizebox{\textwidth}{!}{
 \begin{tikzpicture}\centering
 \begin{scope}[auto, every node/.style={draw,fill=gray!40,circle,inner sep=0, minimum size=2.5em}]
 \node[draw=black,fill=white, minimum size=1em] (A) at (0.25,0.75) {$\tau$};
 \node[draw=none,fill=none, minimum size=0em] (1) at (-0.5,0) {};
 \node[draw=black,fill=white] (2) at (1,0) {$\mathbf{z}^t_{0}$};
 \node[draw=black,fill=white] (3) at (4,0) {$\keywordtri{\mathbf{z}^{t+1}_{0}}$};
  \node[draw=black,fill=white, minimum size=1em] (B) at (4.75,0.75) {{\tiny $L_{t+1}$}};

 \node[draw=black] (4) at (2.5,-2.6) {$\mathbf{z}^t_{T}$};
 \node[draw=none,fill=none, minimum size=0em] (5) at (5.5,0) {};
 \node[draw=none,fill=none, minimum size=0em] (6) at (-0.5,-2.6) {};
 \node[draw=none,fill=none, minimum size=0em] (7) at (1cm-1.25em,-2.6) {};
 \node[draw=none,fill=none, minimum size=2.5em] (71) at (1,-2.6) {};
 \node[draw=none,fill=none, minimum size=2.5em] (8) at (5.5,-2.6) {};
 \path [-, every node/.style={sloped,auto=false}] (1) edge node[align=center] {Clean\\latents} (2);
 \path [->, every node/.style={auto=false}] (A) edge [bend left] node {\tiny \qquad \qquad Eqn.~\eqref{eq:attn}} (2);
 \path [dashed, ->, every node/.style={auto=false}] (3) edge [bend left] node {\tiny Alg.~\ref{alg:extraction} \qquad \qquad} (B);
 \path [-] (2) edge (3);
 \path [->, every node/.style={sloped,auto=false}, keyword] (2) edge [bend left] node[align=center] {interpolating \\ $T$ steps} (3);
 \path [->, every node/.style={sloped,auto=false,anchor=south}] (3) edge node[align=center] {$t$} (5);
 \path [dashed, ->, every node/.style={sloped,auto=false}, Purple] (2) edge [bend right] node[align=center] {sampling \\ $T$ steps} (4);
 \path [->, every node/.style={sloped,auto=false}, Purple] (4) edge [bend right] node[align=center] {denoising \\ $T$ steps} (3);
 \path [-, every node/.style={sloped,auto=false}] (6) edge node[align=center] {Noisy\\latents} (7);
 \path [-] (7) edge (4);
 \path [-] (4) edge (8);
 \path [dashed, ->, every node/.style={sloped,anchor=south,auto=false}, Purple] (3) edge [bend right] (8);
 \end{scope}
 \end{tikzpicture}
 }
 \end{minipage} \hfill
 \begin{minipage}{0.53\textwidth}
 \vspace{-1\baselineskip}
 \renewcommand{\algorithmicindent}{0.5em}%
 \begin{algorithm}[H] %
 \caption{Temporal Interpolation in \textbf{\method}}
 \label{alg:interpolation}
 \begin{algorithmic}[1]
 \INPUT{Network $\keyword{\phi_\theta}$, latent feature $\mathbf{z}^{t}_0$, $\tau \gets \mathcal{T}_\theta(L_0)$}
 \STATE Initialize $\widehat{\mathbf{z}}^{t+1}_T \gets \mathbf{z}^{t}_0$
 \FOR{$k \in \{T, \dots, 0\}$}
 \STATE $\widehat{\mathbf{z}}^{t+1}_k \gets \mathcal{P}_{\keyword{\phi_\theta}}(\widehat{\mathbf{z}}^{t+1}_k, k, \tau)$; \textbf{if $k = 0$ then} break \label{line:network}
 \STATE $\widehat{\mathbf{z}}^{t+1}_{k-1} \gets \widehat{\mathbf{z}}^{t+1}_{k} - \mathcal{Q}(\mathbf{z}^{t}_0, k) + \mathcal{Q}(\mathbf{z}^{t+1}_0, k-1)$ \label{line:offset}
 \ENDFOR %
 \RETURN{$\big\{\widehat{\mathbf{z}}^{t+1}_{k} \; | \; k \in \{T, \dots, 0\}\big\}$}
 \end{algorithmic}
 \end{algorithm}
 \end{minipage}
 \caption{Illustration of the reconstruction and interpolation processes, where the \keywordtwo{purple} dashed arrow is $q(\mathbf{z}_T^t | \mathbf{z}_{0}^t)$ and the \keywordtwo{purple} solid arrow is $p_\varepsilon(\keywordtri{\mathbf{z}_{0}^{t+1}} | \mathbf{z}_{T}^t)$, while the \keyword{blue} arrow illustrates $p_{\keyword{\phi}}(\keywordtri{\mathbf{z}^{t+1}_{0}}|\mathbf{z}^{t}_{0})$.}
 \label{fig:interpolation}
\end{figure}

This \textit{data interpolation model} $\keyword{\phi_\theta}$ (Fig.~\ref{fig:teaser}d) allows us to derive a straightforward \textit{single-step} loss as:
\begin{equation}
 \small
 \mathcal{L} = D_{KL}\big(\keywordtri{\mathbf{z}^{t+1}_{0}} \; \| \; p_{\keyword{\phi}}(\keywordtri{\mathbf{z}^{t+1}_{0}} | \mathbf{z}^{t}_{0})\big) = \log \frac{\keywordtri{\mathbf{z}^{t+1}_{0}}}{p_{\keyword{\phi}}(\keywordtri{\mathbf{z}^{t+1}_{0}} | \mathbf{z}^{t}_{0})}. %
 \label{eq:single_step_loss}
\end{equation}

\begin{table*}[!b]
 \centering
 \caption{Equivalent formulation of interpolative operators, where $\mathbf{z}^{t, t+1}_{k, k-1} = \mathcal{Q}\big(\mathbf{z}^{t, t+1}_0, [k, k-1]\big)$.}
 \resizebox{\textwidth}{!}{
 \setlength\tabcolsep{3pt}
 \begin{tabular}{cccc}
 \toprule
 (a) \textbf{linear blending} & (b) \textbf{learning from $\mathbf{z}^{t+1}_0$} & (c) \textbf{learning from $\mathbf{z}^{t}_0$} & (d) \textbf{learning offset} \\
 \midrule
 $\widehat{\mathbf{z}}^{t+1}_{k-1} = \alpha_{k-1}\ \mathbf{z}^{t}_0~+ $ &
 $\widehat{\mathbf{z}}^{t+1}_{k-1} = \mathbf{z}^{t+1}_0~+ $ &
 $\widehat{\mathbf{z}}^{t+1}_{k-1} = \mathbf{z}^{t}_0~+ $ &
 $\widehat{\mathbf{z}}^{t+1}_{k-1} = \widehat{\mathbf{z}}^{t+1}_{k}~+ $
 \\
 $(1 - \alpha_{k-1})\ \mathbf{z}^{t+1}_0 $ &
 $\frac{\alpha_{k-1}}{\alpha_{k}}\ (\widehat{\mathbf{z}}^{t+1}_{k} - \mathbf{z}^{t+1}_0)$ &
 $\frac{1-\alpha_{k-1}}{1- \alpha_{k}}\ (\widehat{\mathbf{z}}^{t+1}_{k} - \mathbf{z}^{t}_0)$ &
 $(\alpha_k - \alpha_{k-1})(\mathbf{z}^{t+1}_{k-1} - \mathbf{z}^{t}_{k})$
 \\
 \midrule
 \textcolor{OliveGreen}{\textit{stable}} & {\textcolor{BrickRed}{\textit{unstable}}, when $\alpha_k \rightarrow 0$} & {\textcolor{BrickRed}{\textit{unstable}}, when $\alpha_k \rightarrow 1$} & \textcolor{OliveGreen}{\textit{stable}} \\ \midrule
 {\textcolor{OliveGreen}{\textit{deterministic}}} & \textcolor{BrickRed}{\textit{nondeterm.}}, missing $\mathbf{z}^t$ & \textcolor{BrickRed}{\textit{nondeterm.}}, missing $\mathbf{z}^{t+1}$ & \textcolor{OliveGreen}{\textit{deterministic}} \\ \midrule
 {\textcolor{BrickRed}{\textit{nonaccumulative}}} & \textcolor{OliveGreen}{\textit{accumulative}}, Eqn.~\eqref{eq:inductive_process_2b} & \textcolor{OliveGreen}{\textit{accumulative}}, Eqn.~\eqref{eq:inductive_process_2c} & \textcolor{OliveGreen}{\textit{accumulative}}, Eqn.~\eqref{eq:inductive_process} \\ \bottomrule
 \end{tabular}
 }
 \label{tab:equivalent_formulations}
\end{table*}

The simplicity of the loss function comes from the knowledge that we are directly modeling the frame transition in the latent space, that is, $\widehat{\mathbf{z}}^{t+1}_k \approx \mathbf{z}^{t+1}_k$ where $k \in \{T, \dots, 1\}$ is not required. Therefore, we do not use the noise sampling operator $\mathcal{Q}(\cdot)$ as in the step-wise reconstruction objective defined in Eqn.~\eqref{eq:prediction}.
Instead, noise is added in the form of an offset, as described in L\ref{line:offset} of Alg.~\ref{alg:interpolation}. Note that the same network structure of $\varepsilon_\theta$ can be used for $\keyword{\phi_\theta}$ without changing layers. Additionally, with the base case $\widehat{\mathbf{z}}^{t+1}_{T} = \mathbf{z}^{t}_{0}$, the transition is \textit{accumulative} within the \keyword{\textit{inductive}} data interpolation itself:
{\small \begin{align}
\label{eq:inductive_process}
&k \in \{T-1, \dots,1\}, \notag \\
&\Big(\underbrace{\mathcal{P}_{\phi_\theta}\big(\widehat{\mathbf{z}}^{t+1}_{k+1} + (\mathbf{z}^{t+1}_{k} - \mathbf{z}^{t}_{k+1}), k, \tau\big)}_{\keyword{\widehat{\mathbf{z}}^{t+1}_{k}}} \rightarrow \mathcal{P}_{\phi_\theta}\big(\keyword{\widehat{\mathbf{z}}^{t+1}_{k}} \underbrace{+ (\mathbf{z}^{t+1}_{k-1} - \mathbf{z}^{t}_{k})}_{\text{Interpolation operator}}, k - 1, \tau\big)\Big).
\end{align}}

\textbf{Interpolation Operator} is selected based on the theoretical properties between the equivalent variants~\cite{heitz2023iterative}, presented in Table~\ref{tab:equivalent_formulations} and derived in Section~\ref{sec:variants}. In this table, we define $\alpha_k = \frac{k}{T}$, then the selected operator (\ref{tab:equivalent_formulations}d), which adds noise in offset form $\mathcal{Q}(\mathbf{z}^{t+1}_0, k-1) - \mathcal{Q}(\mathbf{z}^{t}_0, k)$, is derived as:
{\small
\begin{align}
 \widehat{\mathbf{z}}^{t+1}_{k-1} = & \hspace{0.4em}\widehat{\mathbf{z}}^{t+1}_{k} + (\alpha_{k} - \alpha_{k-1})\ (\mathbf{z}^{t+1}_{k-1} - \mathbf{z}^{t}_{k}) = \widehat{\mathbf{z}}^{t+1}_{k} + \frac{k - (k -1)}{T}\ (\mathbf{z}^{t+1}_{k-1} - \mathbf{z}^{t}_{k}), \label{eq:scheduling} \\
 \propto & \hspace{0.4em}\widehat{\mathbf{z}}^{t+1}_{k} + (\mathbf{z}^{t+1}_{k-1} - \mathbf{z}^{t}_{k}) = \widehat{\mathbf{z}}^{t+1}_{k} - \mathcal{Q}(\mathbf{z}^{t}_0, k) + \mathcal{Q}(\mathbf{z}^{t+1}_0, k-1), \qquad \text{ as in L\ref{line:offset} of Alg.~\ref{alg:interpolation}}.
\end{align}}

Intuitively, the proposed interpolation process to generate the next frame takes the current frame as the starting point of the noisy sample. The internal states and intermediate features of the diffusion model transition from the current frame, resulting in a more stable prediction for video modeling.

\begin{wrapfigure}[11]{r}{0.57\textwidth}
 \begin{minipage}{0.57\textwidth}
 \vspace{-2\baselineskip}
 \renewcommand{\algorithmicindent}{0.5em}%
 \begin{algorithm}[H] %
 \caption{Correspondence Extraction}
 \label{alg:extraction}
 \begin{algorithmic}[1]
 \INPUT{Internal $Attn$'s while processing $\mathcal{P}_{\keyword{\phi_\theta}}$} %
 \FOR{$k \in [0, T \times 0.8]$}
 \STATE $\mathcal{A}_{S, X} \mathrel{+}= \sum_{l=1}^{N} \big[Attn_{[l,k]}(\varepsilon, \varepsilon), Attn_{[l,k]}(\varepsilon, \tau)\big]$
 \ENDFOR \COMMENT{ requires \textcolor{OliveGreen}{\textit{accumulativeness}} in Table~\ref{tab:equivalent_formulations}}
 \STATE $\bar{\mathcal{A}}_{S, X} \gets \frac{1}{N \times T \times 0.8} \sum_{k=0}^{T\times 0.8} \mathcal{A}_{S, X}$
 \STATE $\bar{\mathcal{A}}^* \gets \bar{\mathcal{A}}_S \circ \bar{\mathcal{A}}_X$
 \STATE $L_{t+1} \gets \mathrm{map}(\bar{\mathcal{A}}^*)$ \COMMENT{ as described in Eqn.~\eqref{eq:extraction}}
 \RETURN{$L_{t+1}$}
 \end{algorithmic}
 \end{algorithm}
 \end{minipage}
\end{wrapfigure}

\noindent\textbf{Correspondence Extraction via Internal States.} From Eqn.~\eqref{eq:attn}, we demonstrate that \textit{the object of interest can be injected via the indication}. From the objectives in Eqn.~\eqref{eq:prediction} and Eqn.~\eqref{eq:interpolation}, we show that \textit{the next frame $\mathbf{I}_{t+1}$ can be \textbf{reconstructed} or \textbf{interpolated} from the current frame $\mathbf{I}_t$}. Subsequently, internal accumulative and stable states, such as the attention map $Attn(\cdot, \cdot)$, which exhibit spatial correlations, can be used to identify the target locations and can be effortlessly extracted. To get into that, the self- and cross-attention maps ($\bar{\mathcal{A}}_{S}$, $\bar{\mathcal{A}}_{X}$) over $N$ layers and $T$ time steps are averaged and performed element-wise multiplication:
\begin{equation}
 \small
 \begin{aligned}
 \bar{\mathcal{A}}_{S} = \frac{1}{N \times T} \sum_{l=1}^{N} \sum_{k=0}^{T} Attn_{[l,k]}(\varepsilon, \varepsilon), \qquad
 \bar{\mathcal{A}}_{X} = \frac{1}{N \times T} \sum_{l=1}^{N} \sum_{k=0}^{T} Attn_{[l,k]}(\varepsilon, \tau), \\
 \bar{\mathcal{A}}^* = \bar{\mathcal{A}}_S \circ \bar{\mathcal{A}}_X, \qquad \bar{\mathcal{A}}^* \in [0, 1]^{H \times W}, \qquad \text{where $(H \times W)$ is the size of }\mathbf{I}_{t + 1}.\label{eq:exponentiation}
 \end{aligned}
\end{equation}

Self-attention captures correlations among latent features, propagating the cross-attention to precise locations. Finally, as in Fig.~\ref{fig:teaser}e, different mappings produce desired prediction types:
\begin{equation}
 \label{eq:extraction}
 \small
 \begin{aligned}
 L_{t + 1} = \mathrm{map}(\bar{\mathcal{A}}^*) =
 \begin{cases}
 \arg \max(\bar{\mathcal{A}}^*), & \text{if point} \\
 \bar{\mathcal{A}}^* > 0, & \text{if segment} \\
 (\min_i\beta, \min_j\beta, \max_i\beta, \max_j\beta), \quad \beta = \big\{(i, j) \; | \; \bar{\mathcal{A}}^*_{i, j} > 0\big\}, & \text{if box}
 \end{cases}
 \end{aligned}
\end{equation}

In summary, the entire diffusion-based tracking process involves the following steps. First, the indication of the object of interest at time $t$ is injected as a condition by $p_\theta(\mathbf{z}^t_0 | \tau)$, derived via Eqn.~\eqref{eq:attn}. Next, the video modeling process operates through the deterministic next-frame interpolation $p_{\keyword{\phi}}(\mathbf{z}^{t+1}_0|\mathbf{z}^t_0)$, as described in Subsection~\ref{method}. Finally, the extraction of the object of interest in the next frame is performed via a so-called ``reversed conditional process'' $p^{-1}_\theta(\mathbf{z}^{t+1}_0 | \tau)$, outlined in Alg.~\ref{alg:extraction}.

\section{Experimental Results}\label{sec:exp}

\subsection{Benchmarks and Metrics}

TAP-Vid~\cite{doersch2022tap} formalizes the problem of long-term physical \textbf{Point Tracking}.
It contains 31,951 points tracked on 1,219 real videos.
Three evaluation metrics are \textit{Occlusion Accuracy (OA)}, $<\delta^{x}_{avg}$ averaging position accuracy, and \textit{Jaccard @ $\delta$} quantifying occlusion and position accuracies.

PoseTrack21~\cite{doering2022posetrack21} is similar to MOT17~\cite{MOT16}. In addition to estimating \textbf{Bounding Box} for each person, the body \textbf{Pose} needs to be estimated. Both keypoint-based and standard MOTA~\cite{bernardin2008evaluating}, IDF1~\cite{ristani2016performance}, and HOTA~\cite{luiten2021hota} evaluate the tracking performance for every keypoint visibility and subject identity.

DAVIS~\cite{perazzi2016benchmark} and MOTS~\cite{porzi2020learning} are included to quantify the \textbf{Segmentation Tracking} performance. For the single-target dataset, evaluation metrics are Jaccard index $\mathcal{J}$, contour
accuracy $\mathcal{F}$ and an overall $\mathcal{J}\&\mathcal{F}$ score~\cite{perazzi2016benchmark}. For the multiple-target dataset, MOTSA and MOTSP~\cite{porzi2020learning} are equivalent to MOTA and MOTP, where the association metric measures the mask IoU instead of the bounding box IoU.

Finally, LaSOT~\cite{fan2021lasot} and GroOT~\cite{nguyen2023type} evaluate the \textbf{Referring Tracking} performance. The \textit{Precision} and \textit{Success} metrics are measured on LaSOT, while GroOT follows the evaluation protocol of MOT.

\subsection{Implementation Details}\label{subsec:details}

We fine-tune the Latent Diffusion Models~\cite{rombach2022high} inplace, follow~\cite{wu2023tune, qi2023fatezero}. However, different from offline fixed batch retraining, our fine-tuning is performed online and auto-regressively between consecutive frames when a new frame is received. Our development builds on LDM~\cite{rombach2022high} for settings with textual prompts and ADM~\cite{DiffusionBeatGANs} for localization settings, initialized by their publicly available pre-trained weights. The model is then fine-tuned using our proposed strategy for 500 steps with a learning rate of $3 \times 10^{-5}$. The model is trained on 4 NVIDIA Tesla A100 GPUs with a batch size of 1, comprising a pair of frames. We average the attention $\bar{\mathcal{A}}_{S}$ and $\bar{\mathcal{A}}_{X}$ in the interval $ k \in [0, T \times 0.8]$ of the DDIM steps with the total timestep $T = 50$.
For the first frame initialization, we employ YOLOX~\cite{ge2021yolox} as the detector, HRNet~\cite{sun2019deep} as the pose estimator, and Mask2Former~\cite{cheng2021per} as the segmentation model. We maintained a linear noise scheduler across all experiments, as it is the default in all available implementations and directly dependent on the number of diffusion steps, which is analyzed in the next subsection. Details for handling multiple objects are in Section~\ref{sec:details}.

\subsection{Ablation Study}\label{subsec:ablation}

\begin{wraptable}[8]{r}{0.6\textwidth}
\vspace{-1.2\baselineskip}
\centering
\caption{The timestep bound $T$ affects reconstruction quality.}
\setlength\tabcolsep{3pt}
\begin{tabular}{c|ccccccc}
\toprule
$T$ (steps) & 50 & 100 & 150 & 200 & 250\\
\midrule
MSE $\downarrow$ & 20.5 & 15.4 & 10.3 & 5.2 & \textbf{0.04} \\
$\mathcal{J}$\&$\mathcal{F}$ $\uparrow$ & 75.4 & 75.8 & 76.0 & 76.3 & \textbf{76.5} \\ \midrule
Reconstruction time (s) $\downarrow$ & 6.2 & 12.7 & 17.5 & 23.6 & 28.7 \\
Interpolation time (s) $\downarrow$ & \textbf{3.2} & 5.7 & 8.5 & 10.6 & 14.7 \\
\bottomrule
\end{tabular}
\label{tab:reconstruction}
\end{wraptable}

\textbf{Diffusion Steps.} We systematically varied the number of diffusion steps (50, 100, 150, 200, 250) and analyzed their impact on performance and efficiency. Results show that we can reconstruct an image close to the origin with a timestep bound $T = 250$ in the reconstruction process of \textbf{\method}.

\begin{table*}[!b]
 \vspace{-1\baselineskip}
 \centering
 \captionsetup{justification=centering}
 \caption{Ablation studies of different temporal modeling alternatives (the second sub-block) and interpolation operators (the third sub-block) on point tracking (A), pose tracking (B), bounding box tracking with and without text (C), and segment tracking (D).}
 \begin{minipage}{.49\textwidth}
 \resizebox{\linewidth}{!}{ %
 \setlength\tabcolsep{3pt}
 \begin{tabular}{l|cc|cc|cc|cc}
 \toprule
 \multirow{2}{*}{\textbf{A. TAP-Vid}} & \multicolumn{2}{c|}{Kinetics} & \multicolumn{2}{c|}{Kubric} & \multicolumn{2}{c|}{DAVIS} & \multicolumn{2}{c}{RGB-Stacking} \\
 & AJ & $<\delta^{x}_{avg}$ & AJ & $<\delta^{x}_{avg}$ & AJ & $<\delta^{x}_{avg}$ & AJ & $<\delta^{x}_{avg}$ \\
 \midrule
 \textbf{DINTR} & \textbf{57.8} & \textbf{72.5} & \textbf{85.5} & \textbf{90.5} & \textbf{62.3} & \textbf{74.6} & \textbf{65.2} & \textbf{77.5} \\ \midrule
 \textit{~(\ref{fig:teaser}c) Recon.} & 53.6 & 64.3 & 80.5 & 86.4 & 62.0 & 66.9 & 62.3 & 71.0 \\ \midrule
 \textit{~(\ref{tab:equivalent_formulations}a) Linear} & 27.6 & 34.8 & 54.6 & 60.1 & 48.1 & 51.6 & 55.6 & 66.3 \\
 \textit{~(\ref{tab:equivalent_formulations}b) $\mathbf{z}_0^{t+1}$} & 34.1 & 43.3 & 64.9 & 63.9 & 51.6 & 54.8 & 59.7 & 60.3 \\
 \textit{~(\ref{tab:equivalent_formulations}c) $\mathbf{z}_0^{t}$} & 33.4 & 41.8 & 63.3 & 62.0 & 51.4 & 53.9 & 58.6 & 59.6 \\ \bottomrule
 \end{tabular}
 }
 \end{minipage} \hfill
 \begin{minipage}{.49\textwidth}
 \resizebox{\linewidth}{!}{ %
 \setlength\tabcolsep{10.2pt}
 \begin{tabular}{l|c|ccccccc}
 \toprule
 \textbf{B. PoseTrack} & \textbf{mAP} & \textbf{MOTA} & \textbf{IDF1} & \textbf{HOTA} \\
 \midrule
 \textbf{DINTR} & \textbf{82.5} & \textbf{64.9} & \textbf{71.5} & \textbf{55.5} \\ \midrule
 \textit{~(\ref{fig:teaser}c) Recon.} & 77.8 & 55.8 & 65.5 & 50.5 \\ \midrule
 \textit{~(\ref{tab:equivalent_formulations}a) Linear} & 59.7 & 39.2 & 43.6 & 34.7 \\
 \textit{~(\ref{tab:equivalent_formulations}b) $\mathbf{z}_0^{t+1}$} & 69.1 & 43.6 & 55.1 & 40.7 \\
 \textit{~(\ref{tab:equivalent_formulations}c) $\mathbf{z}_0^{t}$} & 68.5 & 43.0 & 53.1 & 39.4 \\ \bottomrule
 
 \end{tabular}
 }
 \end{minipage}
 \vspace{\baselineskip}
 
 \begin{minipage}{.49\textwidth}
 \resizebox{\textwidth}{!}{
 \setlength\tabcolsep{3.85pt}
 {\begin{tabular}{l|cc||cc}
 \toprule
 \textbf{C. LaSOT} & \multicolumn{1}{c}{Precision} & \multicolumn{1}{c||}{Success} & \multicolumn{1}{c}{Precision} & \multicolumn{1}{c}{Success} \\ \midrule
 \textbf{DINTR} & \textbf{0.74} & \textbf{0.70} & \textbf{0.60} & \textbf{0.58} \\ \midrule
 \textit{~(\ref{fig:teaser}c) Recon.} & 0.66 & 0.64 & 0.52 & 0.50 \\ \midrule
 \textit{~(\ref{tab:equivalent_formulations}a) Linear} & 0.46 & 0.43 & 0.42 & 0.40 \\ 
 \textit{~(\ref{tab:equivalent_formulations}b) $\mathbf{z}_0^{t+1}$} & 0.52 & 0.49 & 0.46 & 0.45 \\
 \textit{~(\ref{tab:equivalent_formulations}c) $\mathbf{z}_0^{t}$} & 0.51 & 0.48 & 0.44 & 0.44 \\
 \bottomrule
 \end{tabular}}
 }
 \end{minipage} \hfill
 \begin{minipage}{.49\textwidth}
 \resizebox{\textwidth}{!}{
 \setlength\tabcolsep{15pt}
 \begin{tabular}{l|ccc}
 \toprule
 \textbf{D. VOS} & $\mathcal{J}$\&$\mathcal{F}$ & $\mathcal{J}$ & $\mathcal{F}$ \\
 \midrule
 \textbf{DINTR} & \textbf{75.7} & \textbf{72.7} & \textbf{78.6} \\ \midrule
 \textit{~(\ref{fig:teaser}c) Recon.} & 73.9 & 71.8 & 76.1 \\ \midrule
 \textit{~(\ref{tab:equivalent_formulations}a) Linear} & 43.8 & 46.1 & 41.5 \\
 \textit{~(\ref{tab:equivalent_formulations}b) $\mathbf{z}_0^{t+1}$} & 51.1 & 51.3 & 50.9 \\
 \textit{~(\ref{tab:equivalent_formulations}c) $\mathbf{z}_0^{t}$} & 50.5 & 51.0 & 49.9 \\
 \bottomrule
 \end{tabular}
 }
 \end{minipage}
 \label{tab:ab_pose_comparison}
\end{table*}

\textbf{Alternative Approaches} to the proposed \textbf{\method} modeling are discusses in this subsection. To substantiate the discussions, we include all ablation studies in Table~\ref{tab:ab_pose_comparison}, comparing against our base setting. These alternative settings are different interpolation operators as theoretically analyzed in Table~\ref{tab:equivalent_formulations}, and different temporal modeling, including the Reconstruction process as visualized in Fig.~\ref{fig:teaser}c. Results demonstrate that our offset learning approach, which uses two anchor latents to deterministically guide the start and destination points, yields the best performance. This approach provides superior control over the interpolation process, resulting in more accurate and visually coherent output. For point tracking on TAP-Vid, \textbf{\method} achieves the highest scores, with AJ values ranging from 57.8 to 85.5 across different datasets. In pose tracking on PoseTrack, \textbf{\method} scores 82.5 mAP, significantly higher than other methods. For bounding box tracking on LaSOT, \textbf{\method} achieves the highest 0.74 precision and 0.70 success rate with text versus 0.60 precision and 0.58 success rate without text. In segment tracking on VOS, \textbf{\method} scores 75.7 for $\mathcal{J}\&\mathcal{F}$, 72.7 for $\mathcal{J}$, and 78.6 for $\mathcal{F}$, consistently outperforming other methods.

The reconstruction-based method (\ref{fig:teaser}c) generally ranks second in performance across tasks. The decrease in performance for reconstruction is expected, as it does not transfer forward the final prediction to the next step. Instead, it reconstructs everything from raw noise at each step, as visualized in Fig.~\ref{fig:visualize_reconstruction}. Although visual content can be well reconstructed, the lack of seamlessly transferred information between frames results in lower performance and reduced temporal coherence.

The performance difference between (\ref{tab:equivalent_formulations}b) and (\ref{tab:equivalent_formulations}c), which use a single anchor at either the starting latent point ($\mathbf{z}_0^{t}$) or destination latent point ($\mathbf{z}_0^{t+1}$) respectively, is minimal. However, we observed slightly higher effectiveness when controlling the destination point~(\ref{tab:equivalent_formulations}b) compared to the starting point~(\ref{tab:equivalent_formulations}b), suggesting that end-point guidance has a marginally stronger impact on overall interpolation quality. Linear blending~(\ref{tab:equivalent_formulations}a) consistently shows the lowest performance. Derivations of alternative operators blending (\ref{tab:equivalent_formulations}a), learning from $\mathbf{z}_0^{t+1}$ (\ref{tab:equivalent_formulations}b), learning from $\mathbf{z}_0^{t}$ (\ref{tab:equivalent_formulations}c), and learning offset (\ref{tab:equivalent_formulations}d) are theoretically proved to be equivalent as elaborated in Section~\ref{sec:variants}.

\begin{table*}[!t]
 \centering
 \caption{Point tracking performance against several methods on TAP-Vid~\cite{doersch2022tap}.}
 \resizebox{\linewidth}{!}{ %
 \setlength\tabcolsep{4pt}
 \begin{tabular}{l|ccc|ccc|ccc|ccc}
 \toprule
 \multirow{2}{*}{\TAP} & \multicolumn{3}{c|}{Kinetics~\cite{carreira2017quo}} & \multicolumn{3}{c|}{Kubric~\cite{greff2022kubric}} & \multicolumn{3}{c|}{DAVIS~\cite{perazzi2016benchmark}} & \multicolumn{3}{c}{RGB-Stacking~\cite{lee2022beyond}} \\
 & AJ & $<\delta^{x}_{avg}$ & OA & AJ & $<\delta^{x}_{avg}$ & OA & AJ & $<\delta^{x}_{avg}$ & OA & AJ & $<\delta^{x}_{avg}$ & OA \\
 \midrule
 COTR~\cite{jiang2021cotr} & 19.0 & 38.8 & 57.4 & 40.1 & 60.7 & 78.5 & 35.4 & 51.3 & 80.2 & 6.8 & 13.5 & 79.1 \\
 Kubric-VFS-Like~\cite{greff2022kubric} & 40.5 & 59.0 & 80.0 & 51.9 & 69.8 & 84.6 & 33.1 & 48.5 & 79.4 & 57.9 & 72.6 & 91.9 \\
 RAFT~\cite{teed2020raft} & 34.5 & 52.5 & 79.7 & 41.2 & 58.2 & 86.4 & 30.0 & 46.3 & 79.6 & 44.0 & 58.6 & 90.4 \\
 PIPs~\cite{harley2022particle} & 35.1 & 54.8 & 77.1 & 59.1 & 74.8 & 88.6 & 42.0 & 59.4 & 82.1 & 37.3 & 51.0 & 91.6 \\
 TAP-Net~\cite{doersch2022tap} & 46.6 & 60.9 & 85.0 & 65.4 & 77.7 & 93.0 & 38.4 & 53.1 & 82.3 & 59.9 & 72.8 & 90.4 \\
 TAPIR~\cite{doersch2023tapir} & 57.1 & 70.0 & 87.6 & 84.3 & \textbf{91.8} & \textbf{95.8} & 59.8 & 72.3 & 87.6 & \textbf{66.2} & 77.4 & \textbf{93.3} \\ \midrule
 \textbf{\method} & \textbf{57.8} & \textbf{72.5} & \textbf{89.4} & \textbf{85.5} & 90.5 & 95.2 & \textbf{62.3} & \textbf{74.6} & \textbf{88.9} & 65.2 & \textbf{77.5} & 91.6
 \\
 \bottomrule
 \end{tabular}
 }
 \label{tab:sota_point_comparison}
 \vspace{-1\baselineskip}
\end{table*}

\subsection{Comparisons to the State-of-the-Arts}\label{subsec:sota}

\noindent\textbf{Point Tracking.}
As presented in Table~\ref{tab:sota_point_comparison}, our \textbf{\method} point model demonstrates competitive performance compared to prior works due to its thorough capture of local pixels and high-quality reconstruction of global context via the diffusion process. This results in the best performance on DAVIS and Kinetics datasets (88.9 and 89.4 OA).
TAPIR~\cite{doersch2023tapir} extracts features around the estimations rather than the global context.
PIPs~\cite{harley2022particle} and Tap-Net~\cite{doersch2022tap} lose flexibility by dividing the video into fixed segments.
RAFT~\cite{teed2020raft} cannot easily detect occlusions and makes accumulated errors due to per-frame tracking. 
COTR~\cite{jiang2021cotr} struggles with moving objects as it operates on rigid scenes. 

\noindent\textbf{Pose Tracking.} Table~\ref{tab:sota_pose_comparison} compares our \textbf{\method} against other pose-tracking methods.
Classic tracking methods, such as CorrTrack~\cite{rafi2020self} and Tracktor++~\cite{bergmann2019tracking}, form appearance features with limited descriptiveness on keypoint representation.
We also include DiffPose~\cite{feng2023diffpose}, another diffusion-based performer on the specific keypoint estimation task. The primary metric in this setting is the average precision computed for each joint and then averaged over all joints to obtain the final mAP.
DiffPose~\cite{feng2023diffpose} employs a similar diffusion-based generative process but operates on a different heatmap domain, achieving a similar performance on the pixel domain of our interpolation process.

\begin{table*}[!b]
 \vspace{-1\baselineskip}
 \centering
 \begin{minipage}{.495\textwidth}
 \captionsetup{justification=centering}
 \caption{Pose tracking performance against several methods on PoseTrack21~\cite{doering2022posetrack21}.}
 \resizebox{\textwidth}{!}{
 \setlength\tabcolsep{7.7pt}
 \begin{tabular}{l|c|ccccccc}
 \toprule
 \textcolor{OliveGreen}{\textbf{PoseTrack21}} & \textbf{mAP} & \textbf{MOTA} & \textbf{IDF1} & \textbf{HOTA} \\
 \midrule
 CorrTrack~\cite{rafi2020self} & 72.3 & 63.0 & 66.5 & 51.1 \\
 Tracktor++~\cite{bergmann2019tracking} w/ poses & 71.4 & 63.3 & 69.3 & 52.2 \\
 CorrTrack~\cite{rafi2020self} w/ ReID & 72.7 & 63.8 & 66.5 & 52.7 \\
 Tracktor++~\cite{bergmann2019tracking} w/ corr. & 73.6 & 61.6 & 69.3 & 54.1 \\ \midrule
 DCPose~\cite{liu2022temporal} & 80.5 & \xmark & \xmark & \xmark \\
 FAMI-Pose~\cite{liu2021deep} & 81.2 & \xmark & \xmark & \xmark \\
 DiffPose~\cite{feng2023diffpose} & \textbf{83.0} & \xmark & \xmark & \xmark \\ \midrule
 \textbf{\method} & 82.5 & \textbf{64.9} & \textbf{71.5} & \textbf{55.5} \\
 \bottomrule
 \end{tabular}}
 \label{tab:sota_pose_comparison}
 \end{minipage}
 \hfill
 \begin{minipage}{.495\textwidth}
 \captionsetup{justification=centering}
 \caption{Single object tracking without (left) and with (right) textual prompt input.}
 \resizebox{\textwidth}{!}{
 \setlength\tabcolsep{3.75pt}
 \begin{tabular}{l|cc||cc}
 \toprule
 \LSOT & \multicolumn{1}{c}{Precision} & \multicolumn{1}{c||}{Success} & \multicolumn{1}{c}{Precision} & \multicolumn{1}{c}{Success} \\ \midrule
 SiamRPN++~\cite{li2018high} & 0.50 & 0.45 & \xmark & \xmark \\
 GlobalTrack~\cite{huang2020globaltrack} & 0.53 & 0.52 & \xmark & \xmark \\
 OCEAN~\cite{zhang2020ocean} & 0.57 & 0.56 & \xmark & \xmark \\
 \shadetext[left color=blue, right color=OliveGreen, middle color=teal, shading angle=30]{UNICORN}~\cite{yan2022towards} & \textbf{0.74} & 0.68 & \xmark & \xmark \\ \midrule
 GTI~\cite{yang2020grounding} & \xmark & \xmark & 0.47 & 0.47 \\
 AdaSwitcher~\cite{wang2021towards} & \xmark & \xmark & 0.55 & 0.51 \\ \midrule
 \textbf{\method} & \textbf{0.74} & \textbf{0.70} & \textbf{0.60} & \textbf{0.58} \\
 \bottomrule
 \end{tabular}}\label{tab:sot}
 \end{minipage}
\end{table*}

\noindent\textbf{Bounding Box Tracking.} Table~\ref{tab:sot} shows the performance of single object tracking using bounding boxes or textual initialization.
Similarly, Table~\ref{tab:mot} presents the performance of MOT using bounding boxes (left), against DiffussionTrack~\cite{luo2023diffusiontrack} and DiffMOT~\cite{lv2024diffmot} or textual initialization (right), against
MENDER~\cite{nguyen2023type} and MDETR+TrackFormer~\cite{kamath2021mdetr, meinhardt2022trackformer}. Unlike DiffussionTrack~\cite{luo2023diffusiontrack} and DiffMOT~\cite{lv2024diffmot}, which are limited to specific initialization types, our approach allows flexible indicative injection from any type, improving unification capability, and achieving comparable performance. Moreover, capturing global contexts via diffusion mechanics helps our model outperform MENDER and TrackFormer relying solely on spatial contexts formulated via transformer-based learnable queries.

\begin{table*}[!t]
 \caption{Multiple object tracking without (left) and with (right) textual prompt input.}
 \begin{minipage}{.49\textwidth}
 \resizebox{\textwidth}{!}{
 \setlength\tabcolsep{2.6pt}
 {\begin{tabular}{l|cccccccccccc}
 \toprule
 \textcolor{OliveGreen}{\textbf{MOT17}} & \multicolumn{1}{c}{MOTA} & \multicolumn{1}{c}{IDF1} & \multicolumn{1}{c}{HOTA} & \multicolumn{1}{c}{MT} & \multicolumn{1}{c}{ML} & \multicolumn{1}{c}{IDs} \\ \midrule
 MOTR~\cite{zeng2022motr} & 73.4 & 68.6 & 57.8 & 42.9\% & 19.1\% & 2439 \\
 TransMOT~\cite{chu2023transmot} & 76.7 & 75.1 & 61.7 & 51.0\% & 16.4\% & \textbf{2346} \\
 \shadetext[left color=blue, right color=OliveGreen, middle color=teal, shading angle=30]{UNICORN}~\cite{yan2022towards} & 77.2 & 75.5 & 61.7 & \textbf{58.7}\% & \textbf{11.2}\% & 5379 \\
 DiffusionTrack~\cite{luo2023diffusiontrack} & 77.9 & 73.8 & 60.8 & -- & -- & -- \\
 DiffMOT~\cite{lv2024diffmot} & \textbf{79.8} & \textbf{79.3} & \textbf{64.5} & -- & -- & -- \\ \midrule
 \textbf{\method} & 78.0 & 77.6 & 63.5 & 54.2\% & 14.6\% & 4878 \\
 \bottomrule
 \end{tabular}}
 }
 \end{minipage} \hfill
 \begin{minipage}{.517\textwidth}
 \resizebox{\textwidth}{!}{
 \setlength\tabcolsep{4.4pt}
 {\begin{tabular}{l|ccccccccc}
 \toprule
 \GOT & \multicolumn{1}{c}{MOTA} & \multicolumn{1}{c}{IDF1} & \multicolumn{1}{c}{HOTA} & \multicolumn{1}{c}{AssA} & \multicolumn{1}{c}{DetA} \\ \midrule
 MDETR+TFm & 62.6 & 64.7 & 51.5 & 50.9 & 52.2 \\
 MENDER~\cite{nguyen2023type} & 65.5 & 63.4 & 53.2 & 52.9 & 53.7 \\ \midrule
 \textbf{\method} & \textbf{68.9} & \textbf{68.5} & \textbf{57.5} & \textbf{56.9} & \textbf{58.2} \\
 ~(\ref{fig:teaser}c)~\textit{Reconstruct.} & 63.0 & 58.6 & 48.4 & 48.0 & 49.1 \\
 ~(\ref{tab:equivalent_formulations}b)~$\mathbf{z}^{t+1}_0$ & 58.7 & 58.2 & 46.9 & 45.2 & 48.9 \\
 \bottomrule
 \end{tabular}}
 }
 \end{minipage}
 \label{tab:mot}
 \vspace{-1\baselineskip}
\end{table*}

\noindent\textbf{Segment Tracking.} Finally, Table~\ref{table:segment} presents our segment tracking performance against \textit{unified} methods~\cite{wang2021different, yan2022towards}, \textit{single-target} methods~\cite{wang2019fast, voigtlaender2020siam}, and \textit{multiple-target} methods~\cite{voigtlaender2019mots, wu2021track, meinhardt2022trackformer, xu2021segment}. Our \textbf{\method} achieves the best sMOTSA of 67.4, an accurate object tracking and segmentation. Unified methods perform the task separately, either using different branches~\cite{wang2021different} or stages~\cite{yan2022towards}. It leads to a discrepancy in networks. Our \textbf{\method} that is both data- and process-unified avoids this shortcoming.

\begin{table*}[b]
 \vspace{-1\baselineskip}
 \caption{Segment tracking performance on DAVIS~\cite{perazzi2016benchmark} and MOTS~\cite{porzi2020learning}.}
 \begin{minipage}{.42\textwidth}
 \resizebox{\textwidth}{!}{
 \setlength\tabcolsep{5.25pt}
 \begin{tabular}{l|ccc}
 \toprule
 \multirow{1}{*}{\VOS} & $\mathcal{J}$\&$\mathcal{F}$ & $\mathcal{J}$ & $\mathcal{F}$ \\
 \midrule
 SiamMask~\cite{wang2019fast} & 56.4 & 54.3 & 58.5 \\
 Siam R-CNN~\cite{voigtlaender2020siam} & 70.6 & 66.1 & 75.0 \\ \midrule
 \shadetext[left color=blue, right color=OliveGreen, middle color=teal, shading angle=30]{UniTrack}~\cite{wang2021different} & -- & 58.4 & -- \\
 \shadetext[left color=blue, right color=OliveGreen, middle color=teal, shading angle=30]{UNICORN}~\cite{yan2022towards} & 69.2 & 65.2 & 73.2 \\ \midrule
 \textbf{\method} & \textbf{75.4} & \textbf{72.5} & \textbf{78.4} \\
 \bottomrule
 \end{tabular}
 }
 \end{minipage} \hfill
 \begin{minipage}{.57\textwidth}
 \resizebox{\textwidth}{!}{
 \setlength\tabcolsep{4pt}
 {\begin{tabular}{l|cccccccccccc}
 \toprule
 \MOTS & \multicolumn{1}{c}{sMOTSA} & \multicolumn{1}{c}{IDF1} & \multicolumn{1}{c}{MT} & \multicolumn{1}{c}{ML} & \multicolumn{1}{c}{IDSw} \\ \midrule
 Track R-CNN~\cite{voigtlaender2019mots} & 40.6 & 42.4 & 38.7\% & 21.6\% & 567 \\
 TraDeS~\cite{wu2021track} & 50.8 & 58.7 & 49.4\% & 18.3\% & 492 \\
 TrackFormer~\cite{meinhardt2022trackformer} & 54.9 & 63.6 & -- & -- & \textbf{278} \\
 PointTrackV2~\cite{xu2021segment} & 62.3 & 42.9 & 56.7\% & 12.5\% & 541 \\
 \shadetext[left color=blue, right color=OliveGreen, middle color=teal, shading angle=30]{UNICORN}~\cite{yan2022towards} & 65.3 & 65.9 & 64.9\% & 10.1\% & 398 \\ \midrule
 \textbf{\method} & \textbf{67.4} & \textbf{66.4} & \textbf{66.5}\% & \textbf{8.5}\% & 484 \\
 \bottomrule
 \end{tabular}}
 }
 \end{minipage}
 \label{table:segment}
\end{table*}

\section{Conclusion}\label{sec:conclusion}

In conclusion, we have introduced a \textit{Tracking-by-Diffusion} paradigm that reformulates the tracking framework based solely on visual iterative diffusion models.
Unlike the existing denoising process, our \textbf{\method} offers a more seamless and faster approach to model temporal correspondences. This work has paved the way for efficient unified instance temporal modeling, especially object tracking.

\noindent \textbf{Limitations.} There is still a minor gap in performance to methods that incorporate \textit{motion models}, \eg, DiffMOT~\cite{lv2024diffmot} with \textcolor{BrickRed}{\textbf{2D} coordinate} diffusion, as illustrated in Fig.~\ref{fig:teaser}b. However, our novel visual generative approach allows us to handle multiple representations in a unified manner rather than waste $5\times$ efforts on designing specialized models. 
As our approach introduces innovations from \textit{feature representation} perspective, comparisons with advancements stemming from \textit{heuristic optimizations}, such as ByteTrack~\cite{zhang2021bytetrack}, are not head-to-head as these are narrowly tailored increments for a specific type rather than paradigm shifts.
However, exploring integrations between core representation and advancements offers promising performance. Specifically, final predictions are extracted by the so-called ``reversed conditional process'' $p^{-1}_\theta(\mathbf{z}^{t+1}_0 | \tau)$ rather than sophisticated operations~\cite{carion2020end, lin2017feature}.
Finally, time and resource consumption limit the practicality of Reconstruction. However, offline trackers continue to play a vital role in scenarios that demand comprehensive multimodality analysis.

\noindent \textbf{Future Work \& Broader Impacts.} \textbf{\method} is a stepping stone towards more advanced and real-time visual \textit{Tracking-by-Diffusion} in the future, especially to develop a new tracking approach that can manipulate visual contents~\cite{singer2023makeavideo} via the diffusion process or a foundation object tracking model. Specific future directions include formulating diffusion-based tracking approaches for open vocabulary~\cite{li2023ovtrack}, geometric constraints~\cite{dendorfer2022quo}, camera motion~\cite{aharon2022bot, Nguyen_2022_CVPR, nguyen2022multi}, temporal displacement~\cite{zhou2020tracking}, object state~\cite{sun2020simultaneous}, motion modeling~\cite{bewley2016simple, wojke2017simple, lv2024diffmot}, or new object representation~\cite{rajasegaran2022tracking} and management~\cite{stadler2021improving}. The proposed video modeling approach can be exploited for unauthorized surveillance and monitoring, or manipulating instance-based video content that could be used to spread misinformation.

\noindent \textbf{Acknowledgment.} This work is partly supported by NSF Data Science and Data Analytics that are Robust and Trusted (DART), USDA National Institute of Food and Agriculture (NIFA), and Arkansas Biosciences Institute (ABI) grants. We also acknowledge Trong-Thuan Nguyen for invaluable discussions and the Arkansas High-Performance Computing Center (AHPCC) for providing GPUs.

\newpage

%% file: revised_supp.tex
\appendix

{\huge \textbf{Appendices}}

\section{Glossary}\label{appendix}

\begin{table*}[!h]
 \caption{Notations used throughout the paper.}
 \centering
 \setlength\tabcolsep{12pt}
 \renewcommand{\arraystretch}{1.5}
 {\begin{tabular}{p{3cm}l}
 \centering$\mathbf{I}_t$ & Current processing frame (image), $\mathbf{I}_t \in \mathbb{R}^{H\times W\times 3}$\\
 \centering$\mathbf{I}_{t+1}$ & Next frame (image) in the processing video \\
 \centering\multirow{2}{*}{$L_{t}$} & \multicolumn{1}{p{10cm}}{Indicator representation in the current processing frame $\mathbf{I}_t$ (\eg point, bounding box, segment, or \textit{text})} \\
 \centering\multirow{2}{*}{$L_{t+1}$} & \multicolumn{1}{p{10cm}}{Location in the current processing frame $\mathbf{I}_t$ (\eg point, bounding box, or segment)} \\
 \centering$\mathcal{E}(\mathbf{I})$ & Visual encoder $\mathcal{E}$ extracting visual features \\
 \centering$\mathcal{E}(\mathbf{I}_t)[L_{t}]$ & \multicolumn{1}{p{10cm}}{Pooled visual features of the current frame at the indicated location} \\
 \centering$\mathcal{D}(\mathbf{z}_{0})$ & Visual decoder decoding latent feature to image \\
 \centering$\theta$ & Network parameters \\
 \centering$\epsilon$ & A noise variable, $\epsilon \sim \mathcal{N}(0, 1)$ \\
 \centering$\tau$ & Indicator representation \\
 \centering$\varepsilon_\theta(\mathbf{z}_k)$ & Denoising autoencoders, \ie, U-Net blocks \\
 \centering$\phi_\theta(\mathbf{z}_k)$ & Interpolation network, having the same structure as $\varepsilon_\theta$ \\
 \centering$\|\cdot\|_2^2$ & $L^2$ norm \\
 \centering$\mathbf{z}_{0}, \dots, \mathbf{z}_{k}, \dots, \mathbf{z}_{T}$ & Latent variables of the noise sampling process \\
 \centering$\widehat{\mathbf{z}}_{0}, \dots, \widehat{\mathbf{z}}_{k}, \dots, \widehat{\mathbf{z}}_{T}$ & Latent variables of the reconstructive interpolation process \\
 \centering$\alpha_k$ & The scheduling parameter \\
 \centering$\mathcal{Q}(\cdot)$ & Noise sampling process \\
 \centering$\mathcal{P}_{\varepsilon_\theta}(\cdot)$ & Reconstruction/Denoising process, configured by $\varepsilon_\theta$ \\
 \centering$\mathcal{P}_{\phi_\theta}(\cdot)$ & Interpolation process, configured by $\phi_\theta$ \\
 \centering$\mathcal{T}_\theta(\cdot)$ & Indication feature extractor \\
 \centering$\mathbb{E}_{\varepsilon_\theta}L(\cdot)$ & Expectation of a loss function $L(\cdot)$ with respect to $\epsilon_\theta$ \\
 \centering$D_{KL}(P \| Q)$ & Kullback-Leibler divergence of P and Q \\
 \centering$q(\mathbf{z}^t_k | \mathbf{z}^t_{k-1})$ & Conditional probability of $\mathbf{z}^t_k$ given $\mathbf{z}^t_{k-1}$ \\
 \centering$p_\varepsilon(\mathbf{z}^t_{k - 1} | \mathbf{z}^t_{k})$ & Conditional probability of denoising $\mathbf{z}^t_{k - 1}$ given $\mathbf{z}^t_{k}$, configured by $\varepsilon$ \\
 \centering$p_\phi(\widehat{\mathbf{z}}^t_{k - 1} | \widehat{\mathbf{z}}^t_{k})$ & Conditional probability of interpolating $\widehat{\mathbf{z}}^t_{k - 1}$ given $\widehat{\mathbf{z}}^t_{k}$, configured by $\phi$ \\
 \centering$\Big(\mathcal{P}_{\phi_\theta}(\cdot) \rightarrow \mathcal{P}_{\phi_\theta}(\cdot)\Big)$ & Induction process \\
 \centering$\bar{\mathcal{A}}_{S}$ & Average self-attention maps among visual features in U-Net \\
 \centering$\bar{\mathcal{A}}_{X}$ & Average cross-attention maps among visual features in U-Net \\
 \centering$\bar{\mathcal{A}}^{*}$ & Element-wise product of self- and cross-attention \\
 \end{tabular}}
 \label{table:glossary}
\end{table*}

\clearpage
\newpage

\section{Overall Framework}

\begin{figure*}[!t]
 \centering
 \includegraphics[width=1\textwidth]{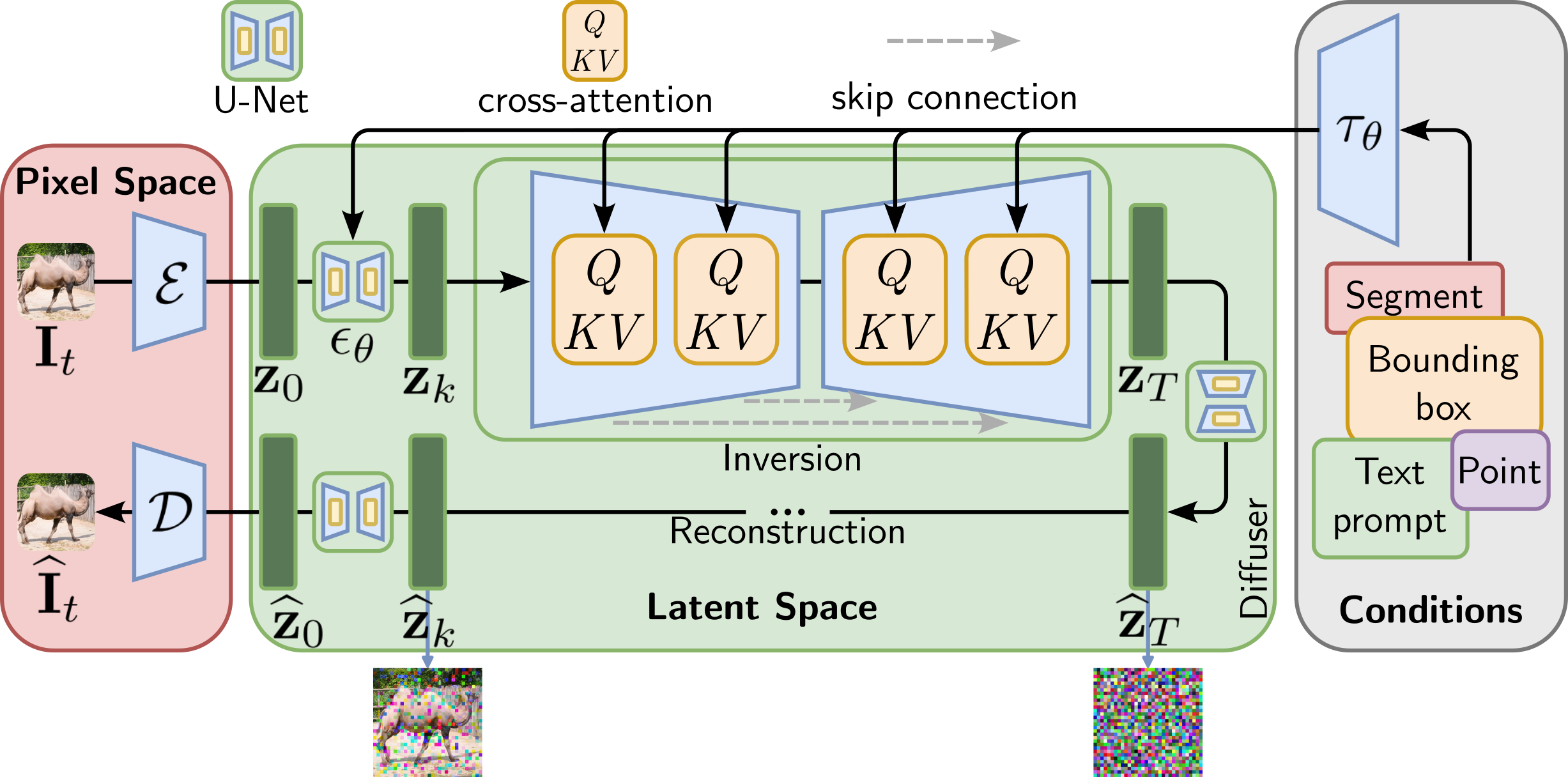}
 \caption{The conditional LDMs utilizes U-Net~\cite{ronneberger2015u} blocks. First, a clean image $\mathbf{I}_k$ is converted to a noisy latent $\mathbf{z}_k$ via the noise sampling process $\mathcal{Q}(\cdot)$ (top branch). Then, well-structured regions are reconstructed from that extremely noisy input via the denoising/reconstruction process $\mathcal{P}_{\varepsilon_\theta}(\cdot)$ (bottom branch). Additionally, conditions can be added as indicators of the regions of interest. While the figure style is adapted from LDMs~\cite{rombach2022high}, we made a distinct change reflecting the \textit{injected} sampling process, following Prompt-to-Prompt~\cite{hertz2022prompt}.
}
 \label{fig:diffusion}
\end{figure*}

\begin{figure*}[!b]
 \centering
 \includegraphics[width=1\textwidth]{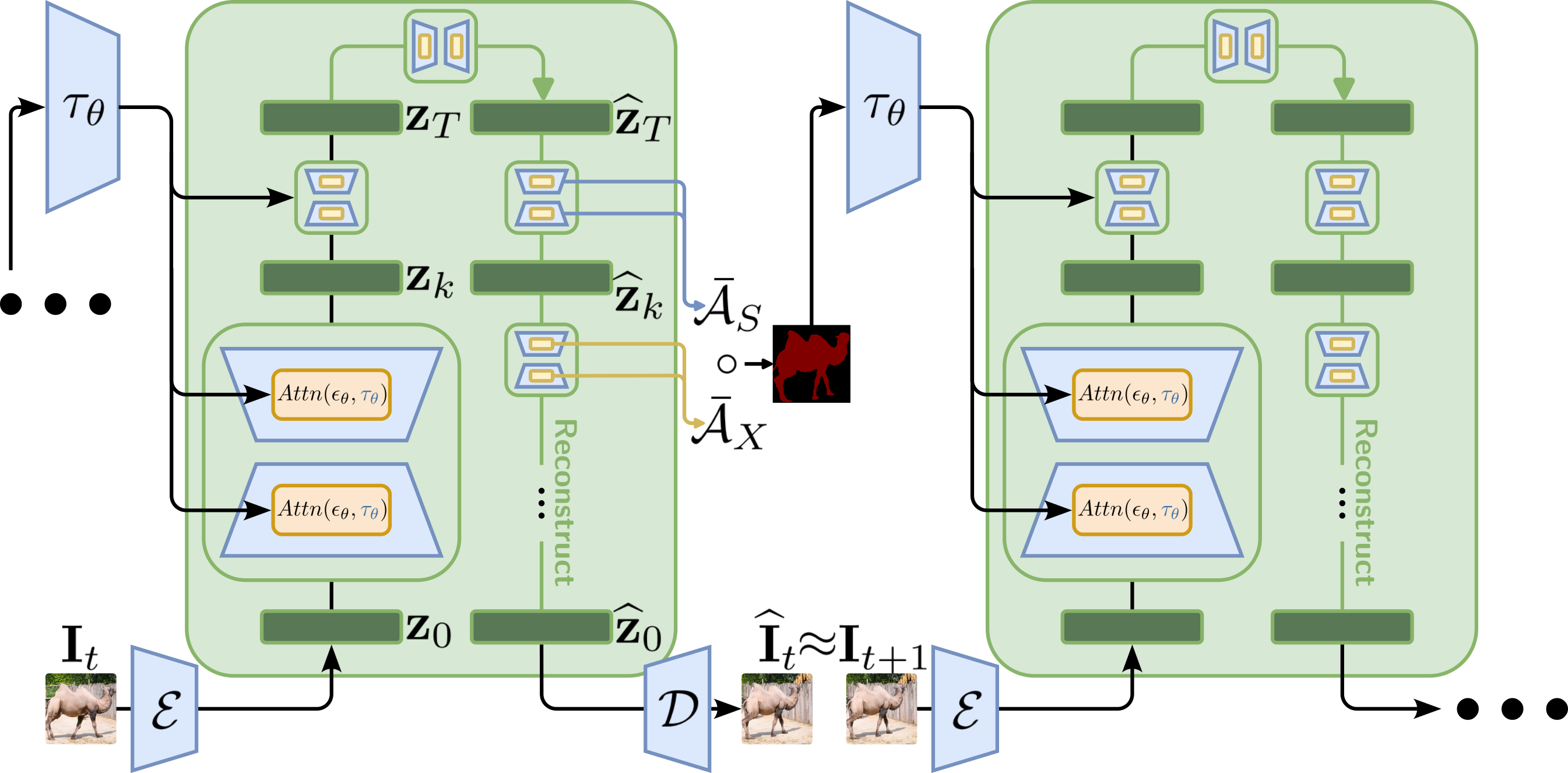}
 \caption{Our proposed autoregressive framework constructed via the diffusion mechanics for temporal modeling. The current frame is input to the encoder $\mathcal{E}(\mathbf{I}_t)$ to produce an initial latent $\mathbf{z}_0$. The sampling process $\mathcal{Q}(\cdot)$ adds noises into the latent in a sequence of $T$ steps. Next, reconstruction process $\mathcal{P}_{\varepsilon_\theta}(\cdot)$ is manipulated through KL divergence optimization \textit{\wrt $\mathbf{z}^{t+1}_{k-1}$}. This shapes the reconstructed image $\widehat{\mathbf{I}}_{t}$ to be more similar to the future frame $\mathbf{I}_{t+1}$. Finally, the location of the targets can be extracted by spatial correspondences, exhibited by the attention maps $\bar{\mathcal{A}}_{S}$ and $\bar{\mathcal{A}}_{X}$.}
 \label{fig:regression}
\end{figure*}

\textbf{Salient Representation.} The ability of the diffuser to, first, \textit{convert a clean image to a noisy latent}, having no recognizable pattern from its origin, and then, \textit{reconstruct well-structured regions from extremely noisy input}, indicates that the diffuser produces powerful semantic contexts~\cite{chen2024your, li2023your}.

In other words, the diffuser can embed semantic alignments, producing coherent predictions between two templates. To leverage this capability, we first consider the generated image $\widehat{\mathbf{I}}_t$ in the diffusion process. Identifying correspondences on the pixel domain can be achieved if:
\begin{equation}\label{correspondences}
 dist\Big(\mathcal{E}(\mathbf{I}_t), \mathcal{E}(\widehat{\mathbf{I}}_{t})\Big) = 0 \text{ is \textit{optimal} from Eqn.~\eqref{eq:diffusion}, then } dist\Big(\mathcal{E}(\mathbf{I}_t)[L_t], \mathcal{E}(\widehat{\mathbf{I}}_{t})[L_t]\Big) = 0.
\end{equation}

We extract the latent features $\mathbf{z}_k$ of their intermediate U-Net blocks at a specific time step $k$ during both processes. This is then utilized to establish injected correspondences between the input image $\mathbf{I}_k$ and the generated image $\widehat{\mathbf{I}}_k$.

\noindent\textbf{Injected Condition.} By incorporating conditional indicators into the Inversion process, we can guide the model to focus on a particular object of interest.
This conditional input, represented as points, poses (\ie, structured set of points), segments, bounding boxes, or even textual prompts, acts as an indicator to \textit{inject the region of interest into the clean latent}, which we want the model to recognize in the reconstructed latent.

These two remarks support the visual diffusion process in capturing and semantically manipulating features for representing and distinguishing objects, as illustrated in Fig.~\ref{fig:diffusion}.
Additionally, Fig.~\ref{fig:regression} presents the autoregressive process that injects and extracts internal states to identify the target regions holding the correspondence temporally.

\section{Derivations of Equivalent Interpolative Operators}\label{sec:variants}

This section derives the variant formulations introduced in Subsection~\ref{method}.

\subsection{Interpolated Samples}

In the field of image processing, an interpolated data point is defined as a weighted combination of known data points through a blending operation controlled by a weighted parameter $\alpha_{k}$:
\begin{align}
\widehat{\mathbf{z}}^{t+1}_{k} &= \alpha_{k} \, \mathbf{z}^{t}_{0} + (1-\alpha_{k}) \, \mathbf{z}^{t+1}_{0}. \label{blending_base}
\end{align}

We can thus rewrite its known samples $\mathbf{z}^{t+1}_{0}$ and $\mathbf{z}^{t}_{0}$ in the following way:
\begin{equation}
\mathbf{z}^{t+1}_{0} = \frac{\widehat{\mathbf{z}}^{t+1}_{k}}{1-\alpha_{k}} - \frac{\alpha_{k} \, \mathbf{z}^{t}_{0}}{1-\alpha_{k}}, \label{eq:appc_x0}
\end{equation}
\begin{equation}
\mathbf{z}^{t}_{0} = \frac{\widehat{\mathbf{z}}^{t+1}_{k}}{\alpha_{k}} - \frac{(1-\alpha_{k}) \, \mathbf{z}^{t+1}_{0}}{\alpha_{k}}. \label{eq:appc_x1}
\end{equation}

\subsection{Linear Blending~(\ref{tab:equivalent_formulations}a)}

In the vanilla version of the algorithm, a blended sample of parameter $\alpha_{k}$ is obtained by blending $\mathbf{z}^{t+1}_{0}$ and $\mathbf{z}^{t}_{0}$, as similar as Eqn.~\eqref{blending_base}:
\begin{align}
\widehat{\mathbf{z}}^{t+1}_{k-1} &= \alpha_{k-1} \, \mathbf{z}^{t}_{0} + (1-\alpha_{k-1}) \, \mathbf{z}^{t+1}_{0}.\label{eq:appc_x_alpha_prime}
\end{align}

To train our interpolation approach using this operator, because the accumulativeness property does not hold, then the step-wise loss as defined in Eqn.~\eqref{eq:data_prediction} has to be employed. As a result, this is equivalent to the reconstruction approach \textit{Reconstruct.} described in Eqn.~\eqref{eq:prediction} and reported in Subsection~\ref{subsec:ablation}.

\subsection{Learning from $\mathbf{z}^{t+1}_{0}$~(\ref{tab:equivalent_formulations}b)}

By expanding $\mathbf{z}^{t}_{0}$ from Eqn.~\eqref{eq:appc_x_alpha_prime} using Eqn.~\eqref{eq:appc_x1}, we obtain:
\begin{align}
\widehat{\mathbf{z}}^{t+1}_{k-1} 
&= (1-\alpha_{k-1}) \, \mathbf{z}^{t+1}_{0} + \alpha_{k-1} \, \mathbf{z}^{t}_{0}, \nonumber\\
&= (1-\alpha_{k-1}) \, \mathbf{z}^{t+1}_{0} + \alpha_{k-1} \, \left(\frac{\widehat{\mathbf{z}}^{t+1}_{k}}{\alpha_k} - \frac{(1-\alpha_k) \, \mathbf{z}^{t+1}_{0}}{\alpha_k}\right), \nonumber\\
&= \left(1- \alpha_{k-1} - \frac{\alpha_{k-1} \,(1-\alpha_k)}{\alpha_k}\right) \, \mathbf{z}^{t+1}_{0} + \frac{\alpha_{k-1}}{\alpha_k} \, \widehat{\mathbf{z}}^{t+1}_{k} ,\nonumber\\
&= \left(\frac{\alpha_k - \alpha_k\, \alpha_{k-1} - \alpha_{k-1} \,(1-\alpha_k)}{\alpha_k}\right) \, \mathbf{z}^{t+1}_{0} + \frac{\alpha_{k-1}}{\alpha_k} \, \widehat{\mathbf{z}}^{t+1}_{k} ,\nonumber\\
&= \left(1- \frac{\alpha_{k-1}}{\alpha_k}\right) \, \mathbf{z}^{t+1}_{0} + \frac{\alpha_{k-1}}{\alpha_k} \, \widehat{\mathbf{z}}^{t+1}_{k} ,\nonumber\\
&= \mathbf{z}^{t+1}_{0} + \frac{\alpha_{k-1}}{\alpha_k} \left(\widehat{\mathbf{z}}^{t+1}_{k} - \mathbf{z}^{t+1}_{0} \right).
\end{align}

\textbf{Inductive Process.} With the base case $\widehat{\mathbf{z}}^{t+1}_{T} = \mathbf{z}^{t}_{0}$, the transition is accumulative within the inductive data interpolation:
\begin{align}
 \label{eq:inductive_process_2b}
 &k \in \{T-1, \dots,1\}, \notag \\
 &\Big(\underbrace{\mathcal{P}_{\phi_\theta}\big(\mathbf{z}^{t+1}_{0} + \frac{\alpha_{k}}{\alpha_{k+1}} (\widehat{\mathbf{z}}^{t+1}_{k+1} - \mathbf{z}^{t+1}_{0} ), k, \tau\big)}_{\keyword{\widehat{\mathbf{z}}^{t+1}_{k}}} \rightarrow \mathcal{P}_{\phi_\theta}\big(\mathbf{z}^{t+1}_{0} + \frac{\alpha_{k-1}}{\alpha_k}(\keyword{\widehat{\mathbf{z}}^{t+1}_{k}} - \mathbf{z}^{t+1}_{0}), k - 1, \tau\big)\Big).
\end{align}

\subsection{Learning from $\mathbf{z}^{t}_{0}$~(\ref{tab:equivalent_formulations}c)}

By expanding $\mathbf{z}^{t+1}_{0}$ from Eqn.~\eqref{eq:appc_x_alpha_prime} using Eqn.~\eqref{eq:appc_x0}, we obtain:
\begin{align}
\widehat{\mathbf{z}}^{t+1}_{k-1} 
&= (1-\alpha_{k-1}) \, \mathbf{z}^{t+1}_{0} + \alpha_{k-1} \, \mathbf{z}^{t}_{0}, \nonumber\\
&= (1-\alpha_{k-1}) \,\left(\frac{\widehat{\mathbf{z}}^{t+1}_{k}}{1-\alpha_k} - \frac{\alpha_k \, \mathbf{z}^{t}_{0}}{1-\alpha_k}\right) + \alpha_{k-1} \, \mathbf{z}^{t}_{0}, \nonumber\\
&= \left( \alpha_{k-1} - \frac{(1-\alpha_{k-1}) \, \alpha_k}{1-\alpha_k}\right) \, \mathbf{z}^{t}_{0} + \frac{1-\alpha_{k-1}}{1-\alpha_k} \, \widehat{\mathbf{z}}^{t+1}_{k} , \nonumber\\
&= \left(\frac{\alpha_{k-1}\, (1- \alpha_k) - (1-\alpha_{k-1}) \, \alpha_k}{1-\alpha_k}\right) \, \mathbf{z}^{t}_{0} + \frac{1-\alpha_{k-1}}{1-\alpha_k} \, \widehat{\mathbf{z}}^{t+1}_{k} , \nonumber\\
&= \left(\frac{1-\alpha_k - (1-\alpha_{k-1})}{1-\alpha_k}\right) \, \mathbf{z}^{t}_{0} + \frac{1-\alpha_{k-1}}{1-\alpha_k} \, \widehat{\mathbf{z}}^{t+1}_{k} , \nonumber\\
&= \left( 1 - \frac{1-\alpha_{k-1}}{1-\alpha_k}\right) \, \mathbf{z}^{t}_{0} + \frac{1-\alpha_{k-1}}{1-\alpha_k} \, \widehat{\mathbf{z}}^{t+1}_{k} , \nonumber\\
&= \mathbf{z}^{t}_{0} + \frac{1-\alpha_{k-1}}{1-\alpha_k} \, \left(\widehat{\mathbf{z}}^{t+1}_{k} - \mathbf{z}^{t}_{0}\right). 
\end{align}

\textbf{Inductive Process.} With the base case $\widehat{\mathbf{z}}^{t+1}_{T} = \mathbf{z}^{t}_{0}$, the transition is accumulative within the inductive data interpolation:
\begin{align}
 \label{eq:inductive_process_2c}
 &k \in \{T-1, \dots,1\}, \notag \\
 &\Big(\underbrace{\mathcal{P}_{\phi_\theta}\big(\mathbf{z}^{t}_{0} + \frac{1-\alpha_{k}}{1-\alpha_{k+1}} \, (\widehat{\mathbf{z}}^{t+1}_{k+1} - \mathbf{z}^{t}_{0}), k, \tau\big)}_{\keyword{\widehat{\mathbf{z}}^{t+1}_{k}}} \rightarrow \mathcal{P}_{\phi_\theta}\big(\mathbf{z}^{t}_{0} + \frac{1-\alpha_{k-1}}{1-\alpha_k} \, (\keyword{\widehat{\mathbf{z}}^{t+1}_{k}} - \mathbf{z}^{t}_{0}), k - 1, \tau\big)\Big).
\end{align}

Due to the absence of the deterministic property and the target term $\mathbf{z}^{t+1}_{0}$, the loss in Eqn.~\eqref{eq:single_step_loss} becomes the sole objective guiding the learning process toward the target. Consequently, we prefer to perform the interpolation operator (\ref{tab:equivalent_formulations}b) in Subsection~\ref{subsec:ablation}, which is theoretically equivalent to this operator.

\subsection{Learning Offset~(\ref{tab:equivalent_formulations}d)}

By rewriting $\alpha_{k-1}=\alpha_{k-1}+\alpha_k-\alpha_k$ in the definition of $\widehat{\mathbf{z}}^{t+1}_{k-1}$, we obtain:
\begin{align}
\widehat{\mathbf{z}}^{t+1}_{k-1} 
&= (1-\alpha_{k-1}) \, \mathbf{z}^{t+1}_{0} + \alpha_{k-1} \, \mathbf{z}^{t}_{0}, \nonumber\\
&= (1-\alpha_{k-1} + \alpha_k-\alpha_k) \, \mathbf{z}^{t+1}_{0} + \left(\alpha_{k-1}+\alpha_k-\alpha_k\right) \, \mathbf{z}^{t}_{0}, \nonumber\\
&= (1-\alpha_k) \, \mathbf{z}^{t+1}_{0} + \alpha_k \, \mathbf{z}^{t}_{0} + \left(\alpha_{k-1}-\alpha_k\right) \, \left(\mathbf{z}^{t}_{0} - \mathbf{z}^{t+1}_{0}\right).
\end{align}

Replace $(1-\alpha_k) \, \mathbf{z}^{t+1}_{0} + \alpha_k \, \mathbf{z}^{t}_{0}$ by $\widehat{\mathbf{z}}^{t+1}_{k}$ from Eqn.~\eqref{blending_base}, we obtain:
\begin{align}
\widehat{\mathbf{z}}^{t+1}_{k-1} 
&= \widehat{\mathbf{z}}^{t+1}_{k} + \left(\alpha_{k-1}-\alpha_k\right) \, \left(\mathbf{z}^{t}_{0} - \mathbf{z}^{t+1}_{0}\right), \nonumber\\
&= \widehat{\mathbf{z}}^{t+1}_{k} + \left(\alpha_k - \alpha_{k-1}\right) \, \left(\mathbf{z}^{t+1}_{0} - \mathbf{z}^{t}_{0}\right), \nonumber\\
&= \widehat{\mathbf{z}}^{t+1}_{k} + \frac{k - (k-1)}{T} \, \left(\mathbf{z}^{t+1}_{0} - \mathbf{z}^{t}_{0}\right).
\end{align}

By multiplying the step $\left(\mathbf{z}^{t+1}_{0} - \mathbf{z}^{t}_{0}\right)$ by a larger factor (\eg, $T$), the scaled step maintain their magnitude and not to become too small when propagated through many layers. Then we obtain:
\begin{align}
\widehat{\mathbf{z}}^{t+1}_{k-1} 
&\propto \widehat{\mathbf{z}}^{t+1}_{k} + \left(\mathbf{z}^{t+1}_{0} - \mathbf{z}^{t}_{0}\right), \quad \text{signified}\\
&\propto \widehat{\mathbf{z}}^{t+1}_{k} + \left(\mathbf{z}^{t+1}_{k-1} - \mathbf{z}^{t}_{k}\right), \label{eq:propto} \\
&= \widehat{\mathbf{z}}^{t+1}_{k} + \Big(\mathcal{Q}\left(\mathbf{z}^{t+1}_{0}, k-1\right) - \mathcal{Q}\left(\mathbf{z}^{t}_{0}, k\right)\Big), \quad \text{as in L\ref{line:offset} of Alg.~\ref{alg:interpolation}}.
\end{align}

\textbf{Inductive Process.} With the base case $\widehat{\mathbf{z}}^{t+1}_{T} = \mathbf{z}^{t}_{0}$, the transition is accumulative within the inductive data interpolation:
\begin{align}
 \label{eq:inductive_process_2d}
 &k \in \{T-1, \dots,1\}, \notag \\
 &\Big(\underbrace{\mathcal{P}_{\phi_\theta}\big(\widehat{\mathbf{z}}^{t+1}_{k+1} + (\mathbf{z}^{t+1}_{k} - \mathbf{z}^{t}_{k+1}), k, \tau\big)}_{\keyword{\widehat{\mathbf{z}}^{t+1}_{k}}} \rightarrow \mathcal{P}_{\phi_\theta}\big(\keyword{\widehat{\mathbf{z}}^{t+1}_{k}} + (\mathbf{z}^{t+1}_{k-1} - \mathbf{z}^{t}_{k}), k - 1, \tau\big)\Big).
\end{align}

\section{Technical Details}\label{sec:details}
\noindent\textbf{Multiple-Target Handling.} Our method processes multiple object tracking by first concatenating all target representations into a joint input tensor during both the Inversion and Reconstruction passes through the diffusion model. Specifically, given $M$ targets, indexed by $i$, each with a indicator representation $L^i_t$, we form the concatenated input:

\begin{equation}
\mathcal{T} = \Big[\mathcal{T}_\theta(L^0_t) \| \dots \| \mathcal{T}_\theta(L^i_t) \| \dots \| \mathcal{T}_\theta(L^{M-1}_t)\Big].
\end{equation}
where $[\ \cdot \ \|\ \cdot\ ]$ is the concatenation operation.

This allows encoding interactions and contexts across all targets simultaneously while passing through the same encoder, decoder modules, and processes. After processing the concatenated output $\mathcal{P}_{\phi_\theta}(\mathbf{z}^{t}_0, T, \mathcal{T})$, we split it back into the individual target attention outputs using their original index order:

\begin{equation}
\bar{\mathcal{A}}_{X} = \Big[\bar{\mathcal{A}}^0_{X} \| \dots \| \bar{\mathcal{A}}^i_{X} \| \dots \| \bar{\mathcal{A}}^{M-1}_{X}\Big], \quad \bar{\mathcal{A}}_{X} \in [0, 1]^{M \times H \times W}.
\end{equation}

So each $\bar{\mathcal{A}}^i_{X}$ contains the refined cross-attention for target $i$ after joint diffusion with the full set of targets. This approach allows the model to enable target-specific decoding. The indices linking inputs to corresponding outputs are crucial for maintaining identity and predictions during the sequence of processing steps.

\noindent\textbf{Textual Prompt Handling.} This setting differs from the other four indicator types, where $L_0$ comes from a dedicated object detector. Instead, we leverage the unique capability of diffusion models to generate from text prompts~\cite{wu2023tune, qi2023fatezero}. Specifically, we initialize $L_0$ using a textual description as the conditioning input. From this textual $L_0$, our process generates an initial set of bounding box proposals as $L_1$. These box proposals then propagate through the subsequent iterative processes to refine into the next $L_2, \dots, L_{|\mathbf{V}| - 2}$ tracking outputs.

\noindent\textbf{Pseudo-code for One-shot Training.} Alg.~\ref{finetuning} and Alg.~\ref{DIFTracker} are the pseudo-code for our fine-tuning and operating algorithms in the proposed approach within the \textit{Tracking-by-Diffusion} paradigm, respectively. The pseudo-code provides an overview of the steps involved in our inplace fine-tuning.

\begin{algorithm}[!h]
\caption{The one-shot fine-tuning pipeline of Reconstruction process}
\begin{algorithmic}[1]
    \INPUT{$\mathbf{I}_t$, $\mathbf{I}_{t+1}$, $\mathcal{T} \gets [\tau_\theta(L^0_t) \| \dots \| \tau_\theta(L^{M-1}_t)]$, $T \gets 50$}
    \STATE $\mathbf{z}_0 \gets \mathcal{E}(\mathbf{I}_t)$
    \STATE $\mathbf{x}_0 \gets \mathcal{E}(\mathbf{I}_{t+1})$
    \STATE $\mathbf{z}_T \gets \mathcal{Q}(\mathbf{z}_0, T)$ \textit{\% injected Inversion}
    \STATE $L_{\text{ELBO}} \gets \mathrm{KL}\big(\mathcal{Q}(\mathbf{x}_{T - 1}, T) \| \mathcal{P}(\mathbf{z}_T, T, \mathcal{T})\big)$ \textit{\% $\ell_T$}
    \FOR{$k \in \{T, \dots, 2\}$}
        \STATE $L_{\text{ELBO}} \mathrel{+}= \mathrm{KL}\big(\mathcal{Q}(\mathbf{x}_{k - 2}, k) \| \mathcal{P}(\widehat{\mathbf{z}}_k, k, \mathcal{T})\big)$ \textit{\% $\ell_{k-1}$}
    \ENDFOR
    \STATE $L_{\text{ELBO}} \mathrel{-}= \log \mathcal{P}(\widehat{\mathbf{z}}_1)$ \textit{\% $\ell_0$}
    \STATE Take gradient descent step on $L_{\text{ELBO}}$
\end{algorithmic}\label{finetuning}
\end{algorithm}

\begin{algorithm}[!h]
\caption{The tracker operation}
\begin{algorithmic}[1]
    \INPUT{Video $\mathbf{V}$, set of tracklets $\mathbf{T} \gets \{L^0_0, \dots, L^{M-1}_0\}$, $\beta = 4$, $T \gets 50$}
    \FOR{$t \in \{0, \dots, |\mathbf{V}| -2\}$}
    \STATE Draw $(\mathbf{I}_t, \mathbf{I}_{t+1}) \in \mathbf{V}$
    \STATE $\mathcal{T} \gets [\tau_\theta(L^0_t) \| \dots \| \tau_\theta(L^{M-1}_t)]$ \textit{\% $\mathcal{T}$ not change if $L^i_t$ is textual prompt}
    \STATE $finetuning(\mathbf{I}_t, \mathbf{I}_{t+1}, \mathcal{T})$ \textit{\% via} Alg.~\ref{finetuning}
    \STATE $\widehat{\mathbf{z}}_{T} \gets \mathcal{P}(\mathbf{z}_{T}, T, \mathcal{T})$
    \FOR{$k \in \{T, \dots, 1\}$}
        \IF{$k \in [1, T \times 0.8]$}
        \STATE $\mathcal{A}_{S} \mathrel{+}= \sum_{l=1}^{N} Attn_{l,k}(\epsilon_\theta, \epsilon_\theta)$
        \STATE $\mathcal{A}_{X} \mathrel{+}= \sum_{l=1}^{N} Attn_{l,k}(\epsilon_\theta, \tau_\theta)$
        \ENDIF
        \STATE $\widehat{\mathbf{z}}_{k} \gets \mathcal{P}(\widehat{\mathbf{z}}_{k+1}, k, \mathcal{T})$
    \ENDFOR
    \STATE $\bar{\mathcal{A}}_{S} \gets \frac{1}{N \times T} \sum_{k=1}^{T} \mathcal{A}_{S}$
    \STATE $\bar{\mathcal{A}}_{X} \gets \frac{1}{N \times T} \sum_{k=1}^{T} \mathcal{A}_{X}$
    \STATE $\bar{\mathcal{A}}^* \gets (\bar{\mathcal{A}}_S)^{\beta} \circ \bar{\mathcal{A}}_X$
    \STATE $[L^0_{t+1} \| \dots \| L^{M-1}_{t+1}] \gets mapping(\bar{\mathcal{A}}^*)$ \textit{\% via} Eqn.~\eqref{eq:extraction}
    \STATE $\mathbf{T} \gets \{L^0_{t+1}, \dots, L^{M-1}_{t+1}\}$
    \ENDFOR
\end{algorithmic}\label{DIFTracker}
\end{algorithm}

\begin{figure*}[!t]
 \centering
 \includegraphics[width=1.0\textwidth]{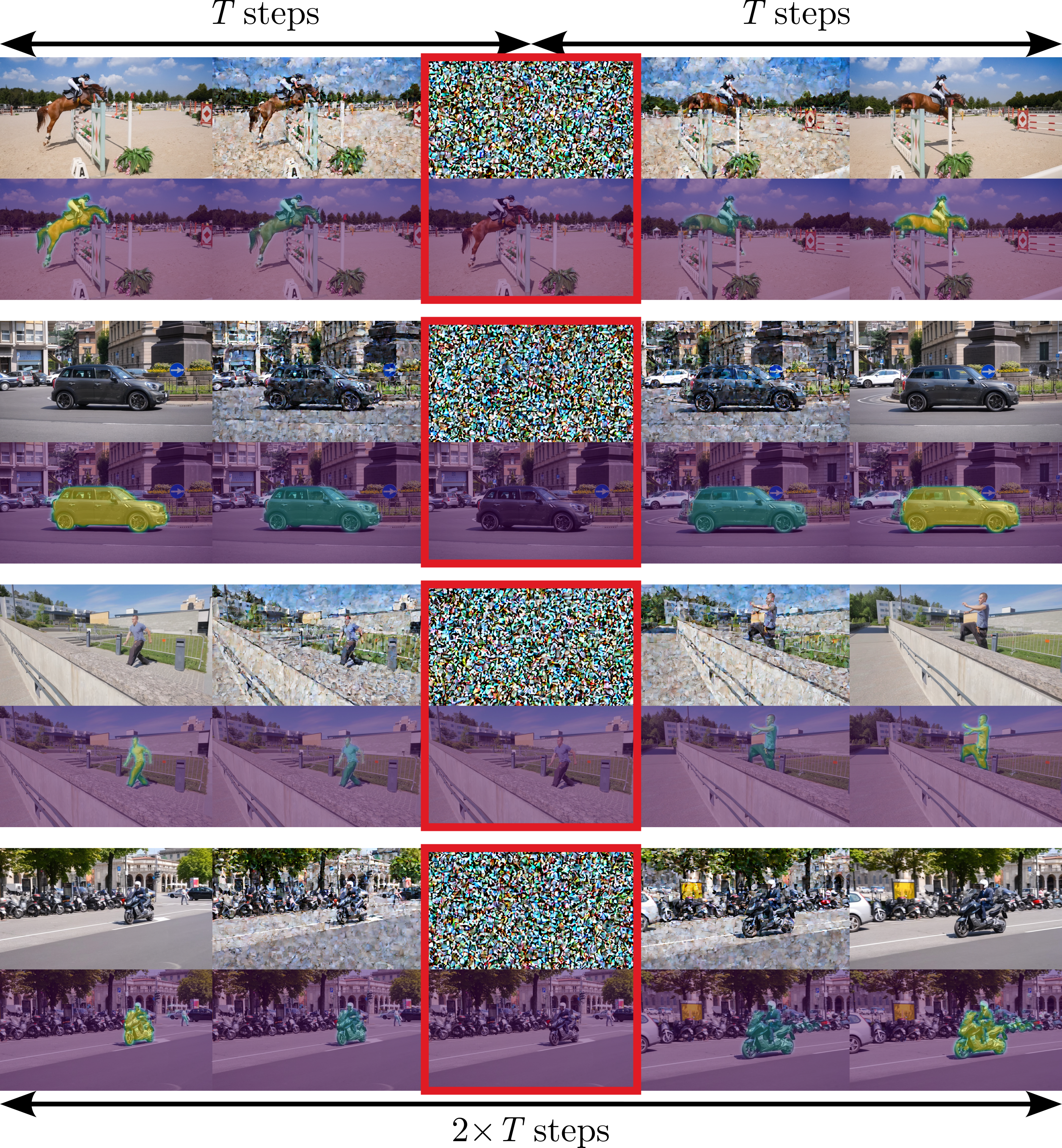}
 \caption{The visualization depicts the diffusion-based Reconstruction process on the DAVIS benchmark~\cite{perazzi2016benchmark}. Unlike the interpolation process in Fig.~\ref{fig:visualize_interpolation}, where internal states are efficiently transferred between frames, the reconstruction process samples visual contents from extreme noise (middle column), and attention maps cannot be transferred. Although visual content can be reconstructed, \textcolor{red}{\textbf{the lack of seamlessly transferred information}} between frames results in lower performance and reduced temporal coherence as in Tables~\ref{tab:sota_point_comparison},~\ref{tab:sota_pose_comparison},~\ref{tab:sot},~\ref{tab:mot}, and~\ref{table:segment}.}
 \label{fig:visualize_reconstruction}
\end{figure*}

\begin{figure*}[!t]
 \centering
 \includegraphics[width=1.0\textwidth]{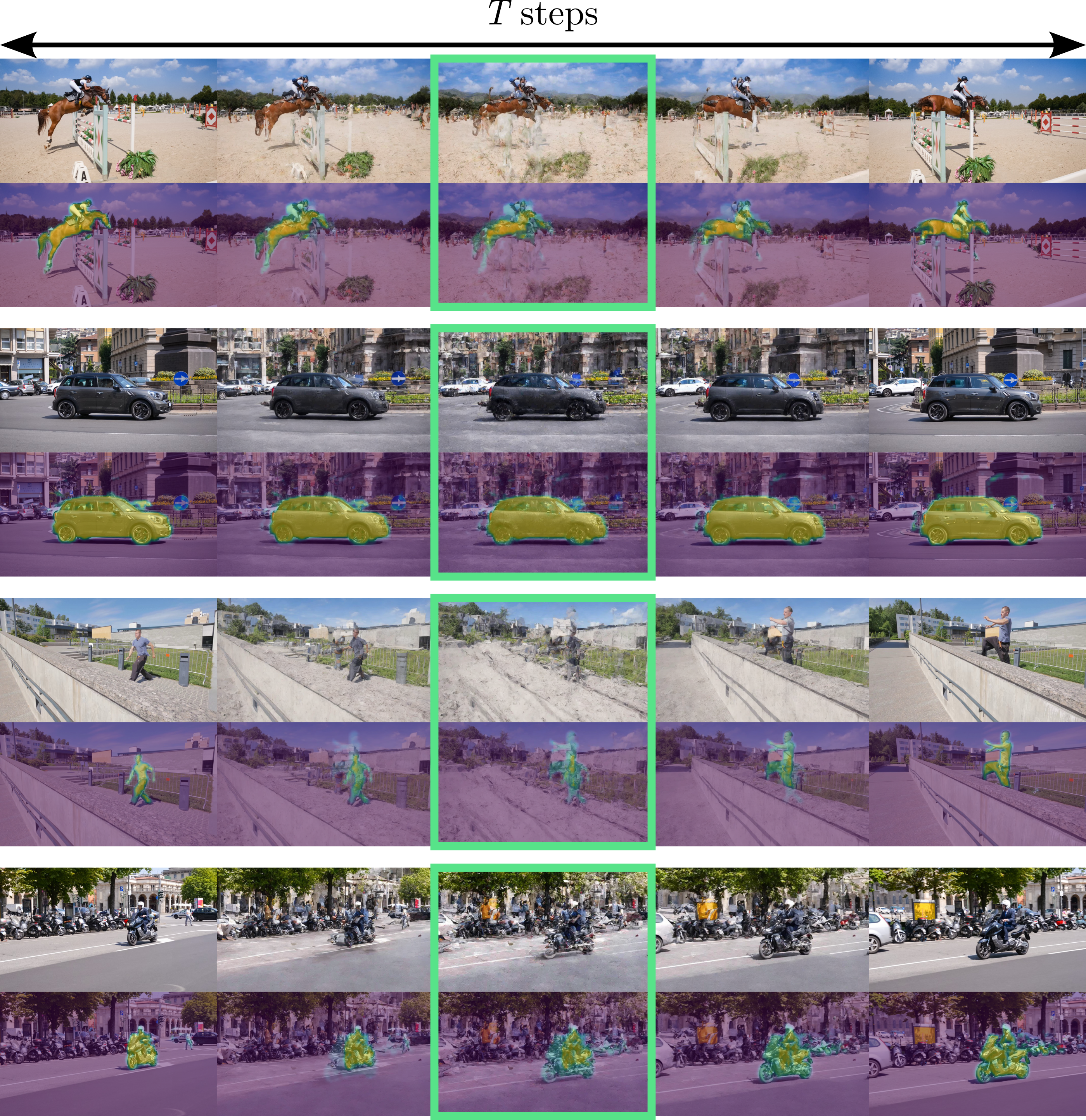}
     \caption{Visualization of the diffusion-based Interpolation process on the DAVIS benchmark~\cite{perazzi2016benchmark}. Different from the reconstruction process in Fig.~\ref{fig:visualize_reconstruction}, where each frame is processed independently, visual contents (top), internal states, and attention maps (bottom) are efficiently transferred from the previous frame to the next frame. This \textcolor{SeaGreen}{\textbf{seamless transfer of information}} between frames results in more consistent and stable tracking, as the model can leverage temporal coherence.}
 \label{fig:visualize_interpolation}
\end{figure*}

\textbf{Process Visualization.} Fig.~\ref{fig:visualize_reconstruction} and Fig.~\ref{fig:visualize_interpolation} are visualizing the two proposed diffusion-based processes that are utilized in our tracker framework.

